\documentclass[lettersize,journal]{IEEEtran}
\usepackage{amsmath,amsfonts}
\usepackage{algorithmic}
\usepackage{algorithm}
\usepackage{booktabs}
\usepackage{enumerate}
\usepackage{bbding}
\usepackage{multirow}
\usepackage{xcolor}
\usepackage{colortbl}
\usepackage[flushleft]{threeparttable}
\definecolor{g1}{HTML}{0eb83a}
\usepackage{array}
\ifCLASSOPTIONcompsoc
    \usepackage[caption=false, font=normalsize, labelfont=sf, textfont=sf]{subfig}
\else
\usepackage[caption=false, font=footnotesize]{subfig}
\usepackage{textcomp}
\usepackage{stfloats}
\usepackage{url}
\usepackage{verbatim}
\usepackage{graphicx}
\usepackage{setspace}
\usepackage{hyperref} 
\usepackage{cite}
\begin{document}

\title{A RankNet-Inspired Surrogate-Assisted Hybrid Metaheuristic\\
for Expensive Coverage Optimization}

\author{Tong-Yu Wu, Chang-Hao Miao, Yun-Tian Zhang, Fang Deng, \emph{Senior Member}, \emph{IEEE}\\ and Chen Chen, \emph{Member}, \emph{IEEE}\thanks{This work was supported by National Natural Science Foundations of China (NSFC) under Grant 62273044; in part by the National Natural Science Foundation of China National Science Fund for Distinguished Young Scholars under Grant 62025301; in part by the National Key Research and Development Program of China No.2022ZD0119703. \emph{(Corresponding authors: Chen Chen)}}
\thanks{Tong-Yu Wu, Chang-Hao Miao, Yun-Tian Zhang, Fang Deng, and Chen Chen are with the National Key Lab of Autonomous Intelligent Unmanned Systems, Beijing Institute of Technology, Beijing 100081, China (e-mail: xiaofan@bit.edu.cn). Tong-Yu Wu and Chang-Hao Miao contributed equally to this work.}}

\markboth{\parbox[t]{\textwidth}{This work has been submitted to the IEEE for possible publication.\\
Copyright may be transferred without notice, after which this version may no longer be accessible.}} %
{How to Use the IEEEtran \LaTeX \ Templates}

\maketitle

\begin{abstract}
Coverage optimization generally involves deploying a set of facilities to best satisfy the demands of specified points, with broad applications in fields such as location science and sensor networks. Recent applications reveal that the subset site selection coupled with continuous angular parameter optimization can be formulated as Mixed-Variable Optimization Problems (MVOPs). Meanwhile, high-fidelity discretization and visibility analysis significantly increase computational cost and complexity, evolving the MVOP into an Expensive Mixed-Variable Optimization Problem (EMVOP). While canonical Evolutionary Algorithms have yielded promising results, their reliance on numerous fitness evaluations is too costly for our problem. Furthermore, most surrogate-assisted methods face limitations due to their reliance on regression-based models. To address these issues, we propose the RankNet-Inspired Surrogate-assisted Hybrid Metaheuristic (RI-SHM), an extension of our previous work. RI-SHM integrates three key components: (1) a RankNet-based pairwise global surrogate that innovatively predicts rankings between pairs of individuals, bypassing the challenges of fitness estimation in discontinuous solution space; (2) a surrogate-assisted local Estimation of Distribution Algorithm that enhances local
exploitation and helps escape from local optima; and (3) a fitness diversity-driven switching strategy that dynamically balances exploration and exploitation. Experiments demonstrate that our algorithm can effectively handle large-scale coverage optimization tasks of up to 300 dimensions and more than 1,800 targets within desirable runtime. Compared to state-of-the-art algorithms for EMVOPs, RI-SHM consistently outperforms them by up to 56.5$\%$ across all tested instances.

\end{abstract}


\section{Introduction}
Coverage optimization deals with mathematical models and algorithms to deploy a set of facilities that satisfy specified customers or demand points in the best possible way. It has an enduring and intense interaction with many other research disciplines, including location science \cite{laporte2019introduction}, sensor networks \cite{ding2021metaheuristics}, geography \cite{ma2023service}, and wireless communication \cite{yuan2024joint}.

In the past few decades, coverage optimization proved its success in wide applications and gained numerous model variants \cite{he2022collaborative}. Driven by modern applications and leveraging structure mathematical optimization, coverage optimization is often formulated as Mixed-Variable Optimization Problems (MVOPs), also known as mixed-integer programming. The discrete variables can represent site selection, which is often described by selecting a subset of candidate sites for deploying facilities (e.g., sensors) \cite{cao2018deployment, saad2020toward}. The continuous variables provide more flexibility in modeling emerging issues in coverage optimization. Take an example: directional sensors (e.g., surveillance cameras, radars) have attracted increasing attention in practical applications and introducing additional complexities such as angular parameter optimization \cite{wu2024mixed, heyns2021optimisation} and power capacity \cite{zhu2023maximal}. MVOPs are challenging due to the multiple disconnected regions in its solution space \cite{liu2023surrogate}. Moreover, the correlation between discrete and continuous variables is a crucial factor affecting overall performance \cite{wu2024mixed}.

\begin{figure}[!t]
    \centering
   \subfloat[Scenario]{%
       \includegraphics[width=0.35\linewidth]{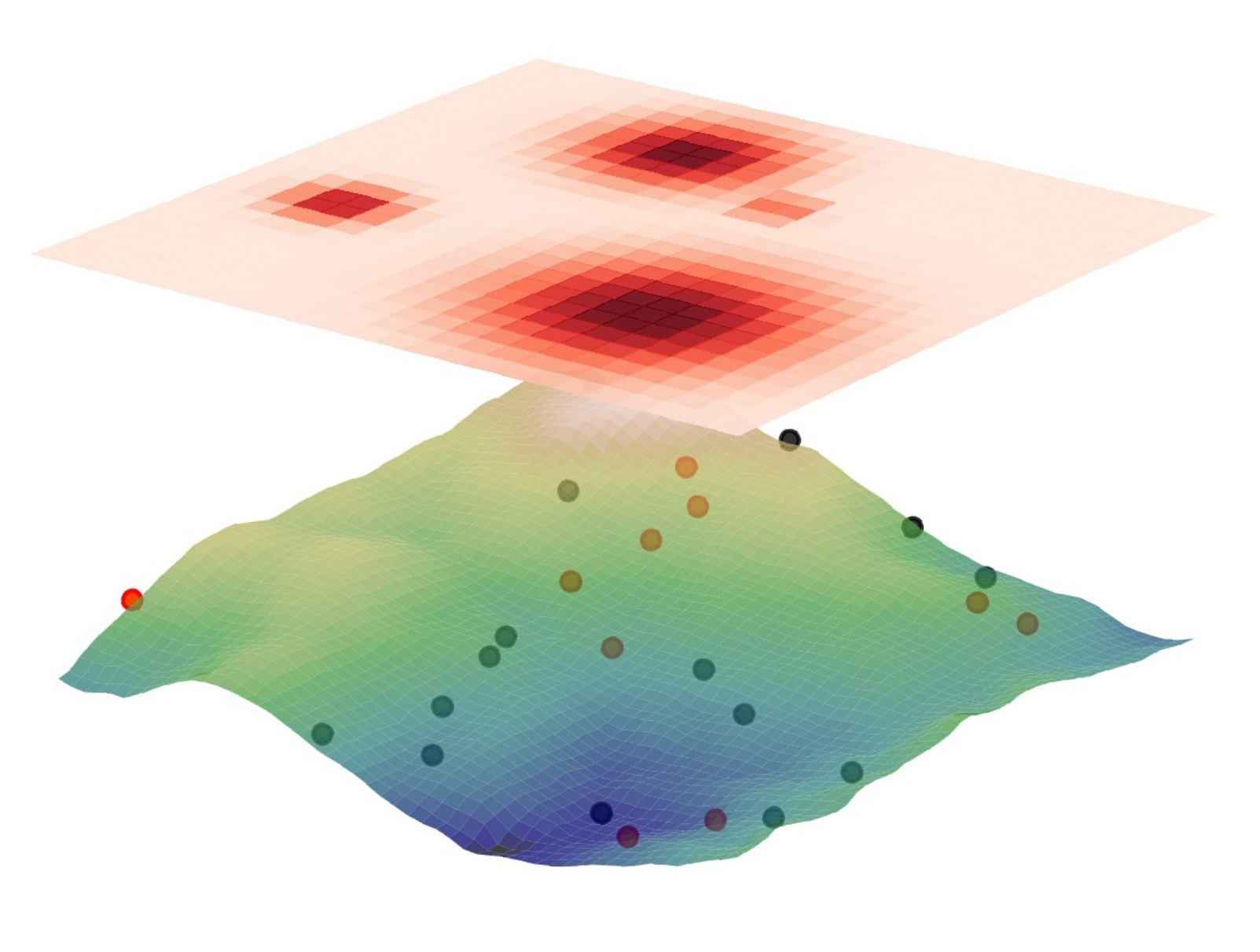}
       \label{fig:coverage_intro}}
    \hfill
    \subfloat[Lower resolution]{%
       \includegraphics[width=0.3\linewidth]{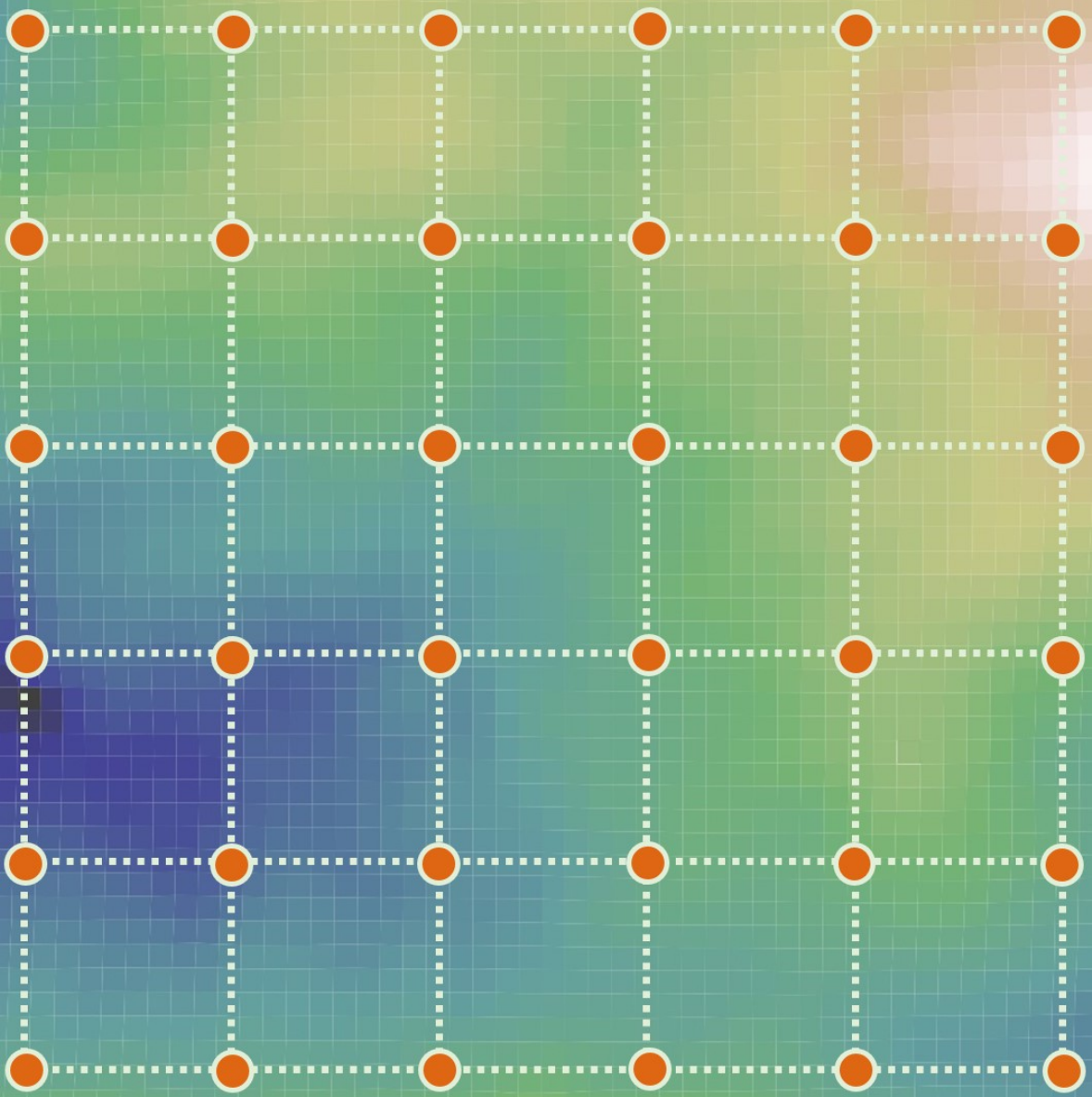}
       \label{fig:Resolution_a}}
    \hfill
    \subfloat[Higher resolution]{%
       \includegraphics[width=0.3\linewidth]{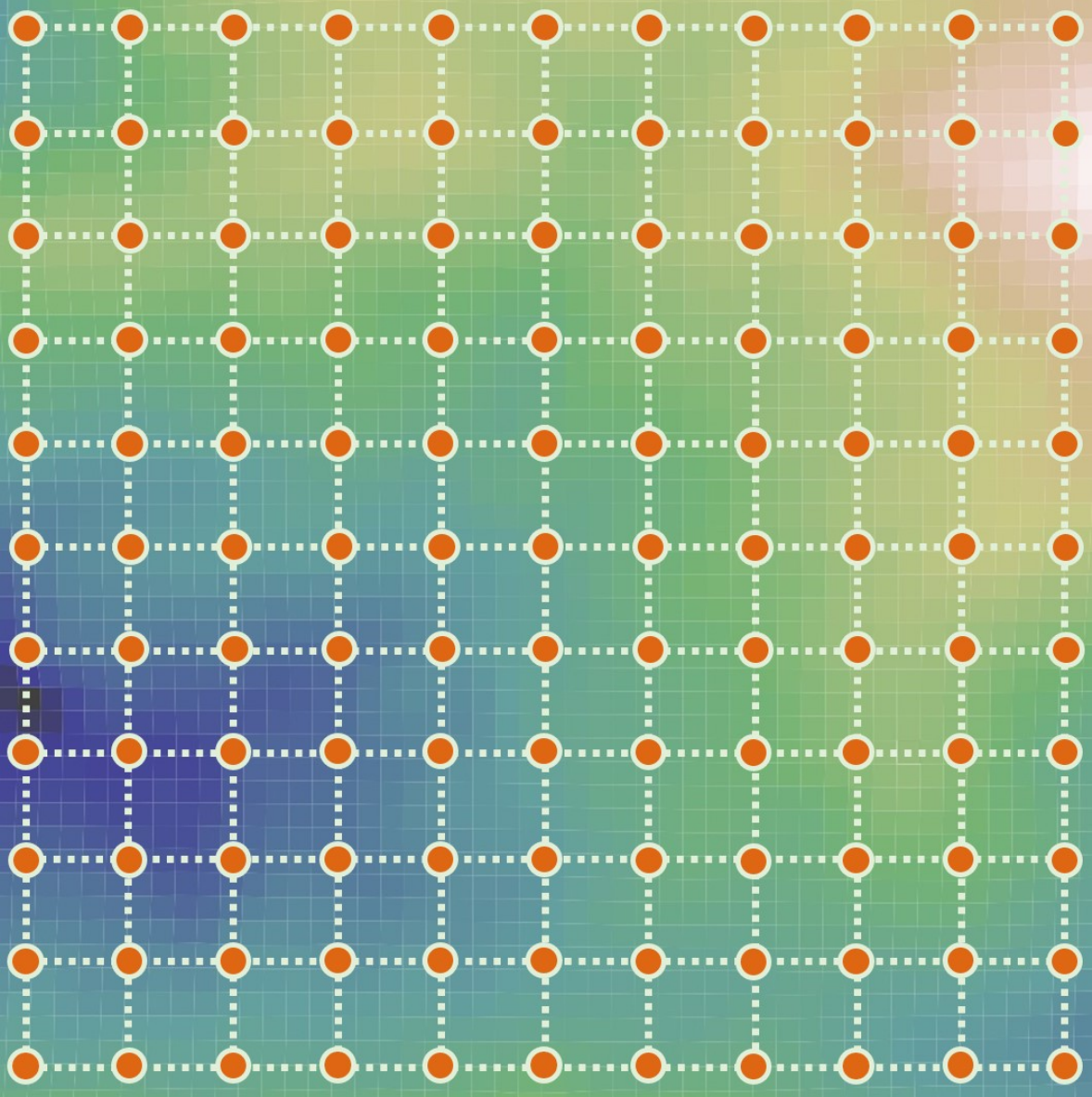}
       \label{fig:Resolution_b}}
    \caption{A high-fidelity realistic coverage optimization scenario in our previous work \cite{wu2024mixed}. The upper heatmap of Fig. \ref{fig:coverage_intro} represents the demand distribution, while Fig. \ref{fig:Resolution_a} and \ref{fig:Resolution_b} show the distribution of demand points within the deployment area under low and high discretization resolutions, respectively. It is obvious that the computational costs are expensive as the pairs of supply-demand points mount up \cite{cordeau2019benders}. For example, solving a case with 100 candidate sites, 300 dimensions, and 30,000 targets requires over 280 hours.}
    \label{fig:intro}
\end{figure}

Moreover, target coverage in coverage optimization can be effectively modeled via the lens of Expensive Optimization Problems (EOPs) due to the following three aspects,
\begin{enumerate}
    \item Target coverage dominates the majority of all literature \cite{he2022collaborative}, and is commonly used as an approximation of comprehensive area coverage, with sampling methods employed to discretize the area \cite{zhang2024acu, heyns2021optimisation, yao2019location}, as shown in Fig. \ref{fig:intro}. Though finer discretization is more realistic, it significantly increases the problem complexity \cite{cordeau2019benders}.
    \item Realistic coverage optimization in 2-D or 3-D space calls for Line-of-Sight (LoS) calculation with visibility analysis. Visibility analysis requires incorporating real Digital Elevation Map (DEM) data and performing calculations via interpolation. As the resolution in 1) increases, the supply and demand point pairs mount, and the Euclidean distance between points grows. Consequently, the computational time required for the problem rises significantly \cite{wrozynski2024reaching}.
    \item Evolutionary Algorithms (EA), also known as metaheuristics, have been widely applied to solve coverage optimization \cite{liao2013ant} and MVOPs due to their robust search capabilities and flexible frameworks \cite{talbi2009metaheuristics}. However, inherited difficulties in 1) and 2) pose significant challenges for individuals' fitness evaluations (FEs) in canonical EAs during evolution.
\end{enumerate}
Thus, high-fidelity realistic coverage optimization can be classified as an Expensive Mixed-Variable Optimization Problem (EMVOP), which calls for efficient solution methods. 

Surrogate-assisted evolutionary algorithms (SAEAs) introduce surrogate models into the EA framework and have been able to efficiently solve EOPs, such as Computational Fluid Dynamics design and Network Architecture Search (NAS)\cite{li2021surrogate}. Therefore, the design of efficient SAEAs has been a key focus in the community. Numerous surrogate models, including Kriging \cite{jones1998efficient}, Radial Basis Function Network (RBFN)\cite{liu2023surrogate}, and Polynomial Response Surface, have been effectively combined with classical EAs \cite{wang2017committee}. These models have demonstrated outstanding performance in solving single and multi-objective optimization problems \cite{song2021kriging}.
%  \cite{jin2018data}
%  \cite{goel2009comparing}

Despite the advancements in SAEAs, research on MVOPs remains less studied. Liu \textit{et al.}\cite{liu2023surrogate} and Xie \textit{et al.} \cite{xie2023dual} introduce global-local SAEA frameworks, which employ RBFN as surrogate models to estimate the best-performing individual for real fitness evaluation. Similarly, Li \textit{et al.} \cite{li2021surrogate} combine the Estimation of Distribution Algorithm (EDA) with Kriging to efficiently solve NAS. Diverging from the evolutionary computation framework, Saves \textit{et al.} \cite{saves2024smt} adopt Kriging as the surrogate model to develop a Bayesian-based algorithm tailored for MVOPs.

However, a common limitation of these methods is their reliance on regression-based surrogate models, which struggle to approximate fitness values accurately. This restricts their applicability to MVOPs with fewer than 100 dimensions. As we know, EAs depend on iterative comparisons between parents and offspring for selection \cite{hao2022expensive}. Due to the discontinuous solution space of MVOP, the performance of regression-based surrogates deteriorates with increasing dimension, as their construction depends on distance-based similarity measures \cite{liu2013gaussian}. Fortunately, pairwise-based approaches can mitigate this issue by converting continuous value predictions into rankings \cite{yuan2021expensive, tian2023pairwise}, which enables the models to infer superior individuals for the next generation without accurate fitness function fitting.

Inspired by recommendation systems \cite{burges2005learning}, we propose the RankNet-inspired Surrogate Hybrid Optimization tailored for realistic coverage optimization. As the extension of our previous work \cite{wu2024mixed}, RI-SHM leverages the well-established global-local SAEA framework, which has proven to be effective in numerous studies \cite{liu2023surrogate, xie2023dual}. The RI-SHM contains three key components: (1) a RankNet-based pairwise global surrogate model, (2) a surrogate-assisted EDA, and (3) an efficient switching strategy.

The main contributions of this paper are summarized as follows:
\begin{enumerate}
    \item We reformulate coverage optimization via the lens of EOPs to create a high-fidelity realistic model and propose an efficient RI-SHM to address this EMVOP.
    \item We construct a RankNet-based pairwise global surrogate to predict the rankings between pairs of individuals. Notably, this represents a novel adaptation of pairwise-based surrogate models for MVOPs implemented within modern deep-learning techniques. In addition, we integrate RBFN with EDA to refine local regions and propose a fitness diversity-driven strategy for adaptively switching between global and local optimization phases.
    \item Experiments demonstrate that our algorithm can effectively handle large-scale coverage optimization tasks of up to 300 dimensions and over 1,800 targets within desirable runtime. Compared to state-of-the-art SAEAs for EMVOPs, our algorithm consistently outperforms them by up to 56.5$\%$ across all tested instances.
\end{enumerate}

The remainder of this paper is organized as follows. Section \ref{sec:related} briefly reviews the background knowledge of this paper. Section \ref{sec:problem} illustrates the high-fidelity realistic model of coverage optimization. The proposed algorithm is described in detail in Section \ref{sec:algo}. Section \ref{sec: experiment} provides the experiment results and analysis. At last, the conclusion and future work are given in Section \ref{sec:conclude}.

\begin{table*}[!t]
\centering
\small
\caption{Summary of common surrogate models in SAEAs}
\label{table:SAEAs}
\begin{threeparttable}
\begin{tabular}{c|ccccc|cc}
\toprule
\textbf{Reference}     & \textbf{Framework} & \textbf{Surrogate models} & \textbf{Category}                 & \textbf{Problems} & \textbf{Scale}  & \textbf{MVOP?}                         & \textbf{Large-scale?}                  \\ \midrule
\cite{liu2013gaussian}          & EC-based  & Kriging          & Regression     & COP      & Medium & -   & -   \\
\cite{zhang2024two}          & EC-based  & Kriging          & Regression     & COP      & Small  & -   & -   \\
\cite{saves2024smt}          & Bayesian  & Mixed-Kriging    & Regression     & MVOP     & Medium & \textcolor{g1}{\CheckmarkBold} & -   \\
\cite{liu2023surrogate, xie2023dual}          & EC-based  & RBFN, Kriging             & Regression     & MVOP     & Large  & \textcolor{g1}{\CheckmarkBold} & \textcolor{g1}{\CheckmarkBold} \\
\cite{lu2014new}          & EC-based  & RankSVM          & Classification & COP      & Small  & -   & -   \\
\cite{zhang2015classification}          & EC-based  & $k$NN              & Classification & MOP      & Small  & -   & -   \\
\cite{pan2018classification}          & EC-based  & DNN              & Classification & MOP      & Small  & -   & -   \\
\cite{dushatskiy2019convolutional}          & EC-based  & CNN              & Pairwise       & BOP      & Large  & -   & \textcolor{g1}{\CheckmarkBold} \\
\cite{yuan2021expensive, hao2022expensive}          & EC-based  & FNN              & Pairwise       & MOP      & Small  & -   & -   \\
\cite{tian2023pairwise}          & EC-based  & PNN              & Pairwise       & MOP      & Small  & -   & -   \\ \midrule
\textbf{Ours} & EC-based  & DNN, RBFN              & Pairwise       & MVOP     & Large  & \textcolor{g1}{\CheckmarkBold} & \textcolor{g1}{\CheckmarkBold} \\ \bottomrule
\end{tabular}
\begin{tablenotes}    
    \item \footnotesize $\bullet$ \textbf{EC-based} stands for evolutionary computation based framework, \textbf{FNN} denotes the feedforward neural network, \textbf{PNN} denotes the probabilistic neural network, and \textbf{COP} denotes continuous optimization problem.
    \item \footnotesize $\bullet$ As noted in \cite{li2021surrogate}, problems with fewer than 30 dimensions are classified as \textbf{small-scale}, those with 30 to 50 dimensions as \textbf{medium-scale}, and those exceeding \textbf{50} dimensions as \textbf{large-scale}.
\end{tablenotes} 
\end{threeparttable}
\end{table*}

\section{Related work}
\label{sec:related}
In this section, we first review the related coverage optimization models for target coverage. Then, we explore recent advancements in surrogate models within SAEAs, focusing primarily on single-objective optimization problems and referencing studies on other problems.

\subsection{Coverage Optimization}
With the rise of sensor networks, coverage optimization has gained increasing attention, which leads to numerous model variants \cite{he2022collaborative}. They can be divided into three main categories: area coverage, target coverage, and barrier coverage \cite{deif2013classification}. Among them, target coverage dominates the majority of the literature, where the area is discretized through sampling methods to approximate comprehensive area coverage \cite{heyns2021optimisation, zhang2024acu, yao2019location}. In this part, we focus on target coverage for sensor networks.

Some researchers model coverage optimization as a location problem in 2-D or 3-D continuous spaces \cite{ding2021metaheuristics}. Due to the practical applications, most studies limit the deployment to finite candidate locations \cite{deif2013classification}, which are typically formulated as discrete site selection tasks. Cao \textit{et al}. \cite{cao2018deployment} consider coverage optimization in 3-D industrial environments, where sensor placement is restricted to candidate edges or vertices of obstacles for safety. Feng \textit{et al}. \cite{feng2021unknown} optimize the localizable \textit{k}-coverage problem in wireless sensor networks to improve connectivity and maximize coverage for industrial Internet of Things (IoT) applications.

Other researchers have approached coverage optimization as MVOPs. These models extend beyond site selection, integrating emerging issues such as power capacity \cite{zhu2023multiobjective} to address real-world constraints. Zhu \textit{et al}. \cite{zhu2023multiobjective} investigate the deployment of visual sensor networks in 3-D space, aiming to maximize target coverage while minimizing power consumption. Specifically, the sensor's working direction and charging period are optimized as continuous variables. Another study by Nguyen \textit{et al}. \cite{nguyen2018efficient} incorporates cluster heads alongside sensor nodes to balance energy consumption for precision farming, which treats energy consumption as continuous variables.

With the increasing adoption of directional sensors in practical applications, we model our coverage optimization as an MVOP that simultaneously optimizes both site selection and angular parameters \cite{saad2020toward, heyns2021optimisation, wu2024mixed}. To ensure variable consistency, previous studies often address the mixed variables by discretizing continuous variables. Saad \textit{et al}. \cite{saad2020toward} aims to maximize target coverage while minimizing the number of sensors. They discretize continuous angles and integrate them with site selection to formulate a binary-encoded Multi-Objective Problem (MOP). Similarly, Heyn \cite{heyns2021optimisation} adopts a coarser approach by discretizing angles into several spatial poses to formulate a discrete optimization problem for backup coverage. However, simplifying discretizing continuous angles may lead to the loss of precision.

To preserve the precision, we proposed $CAGA_{mv}$ that effectively captures the correlation between mixed variables in our previous work \cite{wu2024mixed}. As mentioned before, using target coverage to approximate comprehensive area coverage is commonplace. However, the optimal deployment solution is highly sensitive to the discretization level. While finer discretization provides a more realistic model, it significantly increases the complexity of our problem \cite{cordeau2019benders}. Moreover, realistic coverage optimization requires visibility analysis to obtain LoS. As the pairs of supply and demand points mount up, the computational cost grows substantially \cite{wrozynski2024reaching}. This issue becomes even more severe for population-based EAs \cite{xiang2019clustering}. To address these issues, we introduce SAEAs for more efficient solutions.

\subsection{Surrogate Models in SAEAs}
Surveys on SAEAs have highlighted that model selection, model construction, and model management are critical issues for tackling EOPs \cite{jin2011surrogate, zhou2024evolutionary}. Surrogate models are undeniably central to SAEAs, which significantly influences overall performance \cite{hao2022expensive}. Generally, surrogate models are typically categorized into regression-based, classification-based, and pairwise-based models.

Regression-based models are the most widely used surrogates in SAEAs. They facilitate environment selection by directly predicting the fitness values of individuals. Liu \textit{et al.} \cite{liu2013gaussian} introduce GPEME, a classic SAEA that utilizes Kriging for medium-scale problems, while Zhang \textit{et al.} \cite{zhang2024two} apply Kriging to optimize energy absorption of multi-cell structures in the real world. Moreover, Saves \textit{et al.} \cite{saves2024smt} extend Kriging to a mixed-variable version, enabling its application to MVOPs. However, Kriging's training complexity is $O(N^3\cdot D)$ \cite{liu2013gaussian}, where $N$ is the number of training samples and $D$ stands for the dimension, which becomes computationally prohibitive for large-scale problems. In contrast, RBFNs are favored for their simplicity and lower computational cost \cite{xie2023dual}. Liu \textit{et al.} \cite{liu2023surrogate} propose SHEALED, which employs a global-local framework with RBFNs as surrogates for high-dimensional MVOPs.

Classification-based models often predict the qualitative results between individuals and reference solutions, aligning closely with the selection process in EAs \cite{wei2023hybrid}. Lu \textit{et al.} \cite{lu2014new} employ RankSVM as a classifier to identify the most promising trial vector in Differential Evolution (DE). For MOPs, Zhang \textit{et al.} \cite{zhang2015classification} train a \( k \)-Nearest Neighbor (kNN) classifier to distinguish non-dominated solutions from dominated ones. Similarly, CSEA \cite{pan2018classification} utilizes a Deep Neural Network (DNN) to determine whether individuals dominate predefined reference points.

Recently, pairwise-based models have gained attention. Unlike classification-based surrogates, pairwise-based surrogates focus on determining the relative preference between arbitrary pairs of individuals, eliminating the need for absolute definitions of good or bad solutions \cite{yuan2021expensive}. Dushatskiy \textit{et al.} \cite{dushatskiy2019convolutional} propose the first Convolutional Neural Network (CNN) -based surrogate capable of predicting the fitness relationship between two individuals and validating it on Binary Optimization Problems (BOPs). However, most pairwise-based surrogates are tailored for predicting Pareto dominance \cite{yuan2021expensive, hao2022expensive, tian2023pairwise}, leaving a notable gap in their application to MVOPs. 

Table. \ref{table:SAEAs} briefly concludes the common surrogate models used in SAEAs, highlighting the limited research on efficient algorithms for expensive MVOPs. MVOPs pose significant challenges due to the discontinuity and complexity of the solution space \cite{liu2023surrogate}, making it difficult for regression-based models to predict fitness values accurately. Pairwise-based models offer greater robustness than regression-based ones by focusing on relative ordering rather than precise value predictions. Furthermore, advancements in deep learning provide us with a great opportunity to effectively model mixed-variable correlations, paving the way for efficient solutions to large-scale MVOPs.

\section{Problem Formulation}
\label{sec:problem}
In this section, we describe the formulation of a high-fidelity realistic coverage optimization model aimed at maximizing the coverage of specified targets, which is suggested based on some existing studies \cite{saad2020toward, zhu2023multiobjective, wu2024mixed}. 

\vspace{-0.1cm}
\subsection{Collaborative Coverage Model}
\label{sec:overall_model}
Our coverage optimization involves selecting a subset of $k$ sites from $|Z|$ candidates to deploy directional sensors while determining their corresponding angle parameters. Consequently, each target $q$ is collaboratively covered by a network of multiple sensors, with the coverage probability of $q$ defined as:
\begin{equation}
C(S, q)=1-\prod \limits_{i=1}^k{\left[1-P(s_{i},q)\right]},
\label{target}
\end{equation}
where $P(s_{i},q)$ represents the probability of target point $q$ being covered by sensor $s_{i}$, as detailed in the following subsection.

The sensor network $S$ is required to cover a set of target points $Q$. To facilitate customized surveillance, each target $q\in Q$ is assigned with a weight $\omega_{q}$. Targets closer to critical areas will receive higher weights. We aim to maximize coverage of critical targets while ensuring broad coverage of others. To summarize, we formulate the mathematical model as follows
\begin{equation}
\begin{aligned}
& \underset{b_{j},\theta_{i},\varphi_{i}}{\text{min}}
& & \sum_{q=1}^{|Q|}\omega_{q}\left(\prod \limits_{i=1}^k{\left[1-P(s_{i},q)\right]}\right) \\
& \text{subject to}
& & \sum_{j=1}^{|Z|}b_{j}=k, \; b_{j} \in \{0,1\}.\\
& & & \theta_{i} \in [-180^{\circ},180^{\circ}], \; i = 1, \ldots, k.\\
& & & \varphi_{i} \in [-90^{\circ},90^{\circ}], \; i = 1, \ldots, k.
\end{aligned}
\label{eq:model}
\end{equation}
where $b_{j}$ is a binary variable indicating whether site $j$ is selected, and $\theta_{i},\varphi_{i}$ represent the pan and tilt angles of sensor $s_{i}$, respectively.

The proposed model incorporates two types of variables: discrete variables $b_{j}$ and continuous variables $\theta_{i}, \varphi_{i}$. Consequently, this results in a mixed-variable programming model that is both non-convex and nonlinear. Moreover, the inherent correlation between the discrete and continuous variables introduces additional complexity that cannot be overlooked. We have made preliminary attempts to address this issue in our previous work \cite{wu2024mixed}.

\vspace{-0.12cm}
\subsection{Probabilistic Sensing Model}
Recent research has shifted from traditional binary and omnidirectional models to directional and probabilistic ones, which better address real-world uncertainties. In 3-D space, terrain obstructions further constrain coverage, which requires visibility analysis that often involves costly interpolation. Formally, the coverage probability is defined as,
\begin{equation}
P(s_{i},q)=\mu_{d}(||s_{i}-q||)\times \mu_{p}(\alpha_{p_{qi}}) \times \mu_{t}(\alpha_{t_{qi}}) \times v(s_{i},q).
\label{coverage}
\end{equation}

The distance member function $\mu_{d}(||s_{i}-q||)$ provides a coverage metric based on the Euclidean distance between the sensor $s_{i}$ and the target $q$.
\begin{equation}
\mu_{d}(||s_{i}-q||)=1-\frac{1}{1+exp(-\beta_{d}(||s_{i}-q||-t_{d}))},
\label{distance member}
\end{equation}
where $||s_{i}-q||=\sqrt{\Delta x^2 + \Delta y^2 + \Delta z^2}$, $\Delta x = x_{q}-x_{i}, \Delta y = y_{q}-y_{i}, \Delta z = z_{q}-z_{i}$. $(x_{i}, y_{i}, z_{i})$ and $(x_{q}, y_{q}, z_{q})$ are the coordinates of sensor $s_{i}$ and target $q$, respectively.

The pan member function $\mu_{p}(\alpha_{p_{qi}})$ and tilt member function $\mu_{t}(\alpha_{t_{qi}})$ evaluate coverage based on the horizontal deviation $\alpha_{p_{qi}} \in [-180^{\circ},180^{\circ}]$ and vertical deviation $\alpha_{t_{qi}} \in [-90^{\circ},90^{\circ}]$ from the direction of sensor $s_{i}$, respectively.
\begin{equation}
\begin{aligned}
\mu_{p}(\alpha_{p_{qi}})&=\frac{1}{1+exp(-\beta_{p}(\alpha_{p_{qi}}+t_{p}))}\\
&-\frac{1}{1+exp(-\beta_{p}(\alpha_{p_{qi}}-t_{p}))},
\end{aligned}
\label{pan member}
\end{equation}
where $\alpha_{p_{qi}}=arccos(\frac{cos\theta_{i}\Delta x+sin\theta_{i}\Delta y}{\sqrt{\Delta x^2 + \Delta y^2}})$, $\theta_{i}$ is the pan angle of the sensor with range of $[-180^{\circ},180^{\circ}]$.

\begin{equation}
\begin{aligned}
\mu_{t}(\alpha_{t_{qi}})&=\frac{1}{1+exp(-\beta_{t}(\alpha_{t_{qi}}+t_{t}))}\\
&-\frac{1}{1+exp(-\beta_{t}(\alpha_{t_{qi}}-t_{t}))},
\end{aligned}
\label{tilt member}
\end{equation}
where $\alpha_{t_{qi}}=arctan(\frac{\Delta z}{\sqrt{\Delta x^2 + \Delta y^2}})-\varphi_{i}$, $\varphi_{i}$ is the tilt angle of the sensor with range of $[-90^{\circ},90^{\circ}]$.

Coverage of target $q$ by sensor $s_{i}$ requires $q$ to lie within $s_{i}$'s detection range; in addition, the visibility (i.e., Line-of-Sight, LoS) between them should be ensured. We employ the Bresenham LoS algorithm \cite{temel2013deployment} to analyze the visibility between two points and represent the presence of visibility using a binary variable $v(s_{i},q)\in\{0,1\}$. Although the Bresenham LoS algorithm avoids extensive interpolation calculations, it becomes computationally expensive in high-dimensional, large-scale scenarios. For instance, solving a problem with 100 candidate sites, 300 dimensions, and 30,000 targets requires over 280 hours.

\subsection{Discussion}
As shown in Eq. \ref{eq:model}, the objective function of our coverage optimization minimizes the weighted sum of coverage blind spots for all target points $Q$. In our problem, we consider target coverage as an approximation of comprehensive area coverage, with sampling methods employed to discretize the deployment area. The finer the discretization, the harder it is to solve the resulting problem \cite{cordeau2019benders}. 

Furthermore, Eq. \ref{coverage} indicates that realistic coverage optimization requires visibility analysis $v(s_{i},q)$, which relies on real DEM data. As the resolution of discretization increases, both the number of supply-demand point pairs $(s_{i},q)$ and the Euclidean distance between points grow. As a result, the corresponding computational cost arises significantly \cite{wrozynski2024reaching}. This explains why high-fidelity realistic coverage optimization evolves into EMVOP challenges.

The issues outlined above pose a significant challenge for canonical EAs, as the evaluation of each individual's fitness dramatically increases the computational burden \cite{jin2011surrogate}. To address this, we propose a novel SAEA inspired by RankNet to enhance evolutionary efficiency in the next section.

\begin{figure*}[!t]
    \centering
    \includegraphics[width=1.0\textwidth]{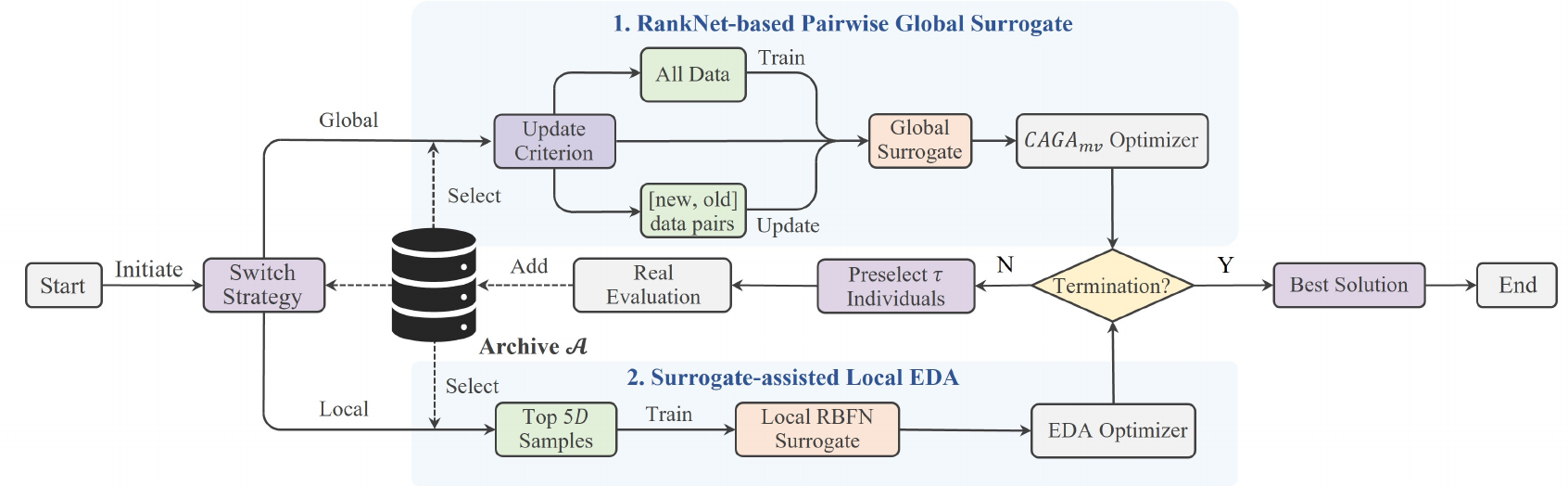}
    \caption{The generic framework of the proposed RI-SHM consists of three key components: a RankNet-based pairwise global surrogate model, a surrogate-assisted local EDA, and a fitness diversity-driven switching strategy.}
    \label{fig:framework}
\end{figure*}

\section{Proposed algorithm}
\label{sec:algo}
\subsection{Overall Framework}
The high-fidelity realistic coverage optimization of sensor networks, classified as an EMVOP, can benefit from surrogate-assisted optimization to enhance efficiency and performance. To this end, we introduce the RI-SHM within a global-local SAEA framework, which consists of three key components: a RankNet-based pairwise global surrogate, a surrogate-assisted local EDA, and a fitness diversity-driven switching strategy. The generic framework of our RI-SHM is shown in Fig. \ref{fig:framework}, which mainly includes the following five steps:
\subsubsection{Initialization} To ensure a uniform exploration of the mixed-variable solution space, we use Sobol sampling for discrete variables and Latin Hypercube Sampling for continuous ones, respectively. A total of $2D$ samples are generated and evaluated with a real fitness function to initialize the data archive $\mathcal{A}$, where $D$ indicates the dimension of decision variables. The top-ranked $N$ individuals are then selected to form the initial population \cite{wang2019novel}, where $N$ is the population size.

\subsubsection{Fitness Diversity-Driven Switching Strategy} We employ the fitness diversity metric to guide the switching between global and local optimization. When current diversity falls below the threshold $\delta$, the alternative strategy is activated to discover diverse genotypes.

\subsubsection{RankNet-Based Pairwise Global Surrogate} For global optimization, we propose a novel global surrogate based on the RankNet framework \cite{burges2005learning}, which employs pairwise training to rank individuals through parent-offspring comparison. Initially, the global surrogate is trained using all data from the archive $\mathcal{A}$. To reduce the computational cost, the model is updated only when over $T$ new individuals are added to $\mathcal{A}$ and the global optimum improves. These new individuals are paired with old ones to generate data pairs for model updating; otherwise, the existing model is used for inference. Additionally, we design a data augmentation technique to enhance generalization for our deployment problem. For optimization, we adopt our proposed $CAGA_{mv}$ \cite{wu2024mixed} to generate new individuals. The surrogate preselects $\tau$ individuals for real fitness evaluation, which are then added to $\mathcal{A}$.

% orr1996introduction
\subsubsection{Surrogate-Assisted Local Estimation of Distribution} For local optimization, a distance-based RBFN is constructed as the local surrogate, while the weighted EDA \cite{li2021surrogate} serves as the local optimizer. The RBFN is trained with the top-ranked $5D$ samples from the archive $\mathcal{A}$ \cite{liu2023surrogate}. After weighted EDA samples offspring, the best predicted and most uncertain solutions are preselected for real fitness evaluation and added to $\mathcal{A}$ .

\subsubsection{Termination} When the number of real FEs exceeds $MaxFEs$, the algorithm outputs the best solution during the evolution. Otherwise, RI-SHM goes back to Step 2 and iteratively repeats the procedure.

\subsection{RankNet-based Pairwise Global Surrogate}
In the global-local surrogate-assisted optimization framework, the global surrogate and its optimizer should effectively explore the high-dimensional mixed-variable solution space to identify promising regions. Building on our previous work \cite{wu2024mixed}, we adopt $CAGA_{mv}$, a variant of GA, as the global optimizer due to its effective correlation-aware mechanism. The iteration of GA relies on the selection through parent-offspring comparison, which aligns well with pairwise-based surrogate models. It drives population evolution by predicting the rankings between pairs of individuals. Compared to regression-based surrogates, pairwise-based models are more robust, focusing on relative ordering rather than precise value prediction.
%pairs of individuals \cite{naharro2022comparative}

Similarly, researchers score and rank candidate items based on user preferences in recommendation systems to generate personalized recommendation lists. RankNet is a classic ranking framework that uses neural networks to model ranking functions, converting an ordinal regression problem into a more straightforward binary classification task \cite{burges2005learning}. For example, if item $A$ is ranked higher than $B$, and $B$ is ranked higher than $C$, then $A$ must be ranked higher than $C$ to maintain consistency \cite{bradley1952rank}. To enforce this condition, RankNet applies the \textit{sigmoid} function to calculate the preference probability, which is based on the difference between the network outputs of input pairs.
% as follows:
% \begin{equation}
%     P_{AB} = \frac{e^{O_{AB}}}{1 + e^{O_{AB}}},
% \label{sigmoid}
% \end{equation}
% where $O_{AB}$ is the difference between the network outputs of $A$ and $B$.   

\begin{figure}[!t]
    \centering
    \includegraphics[width=1.0\linewidth]{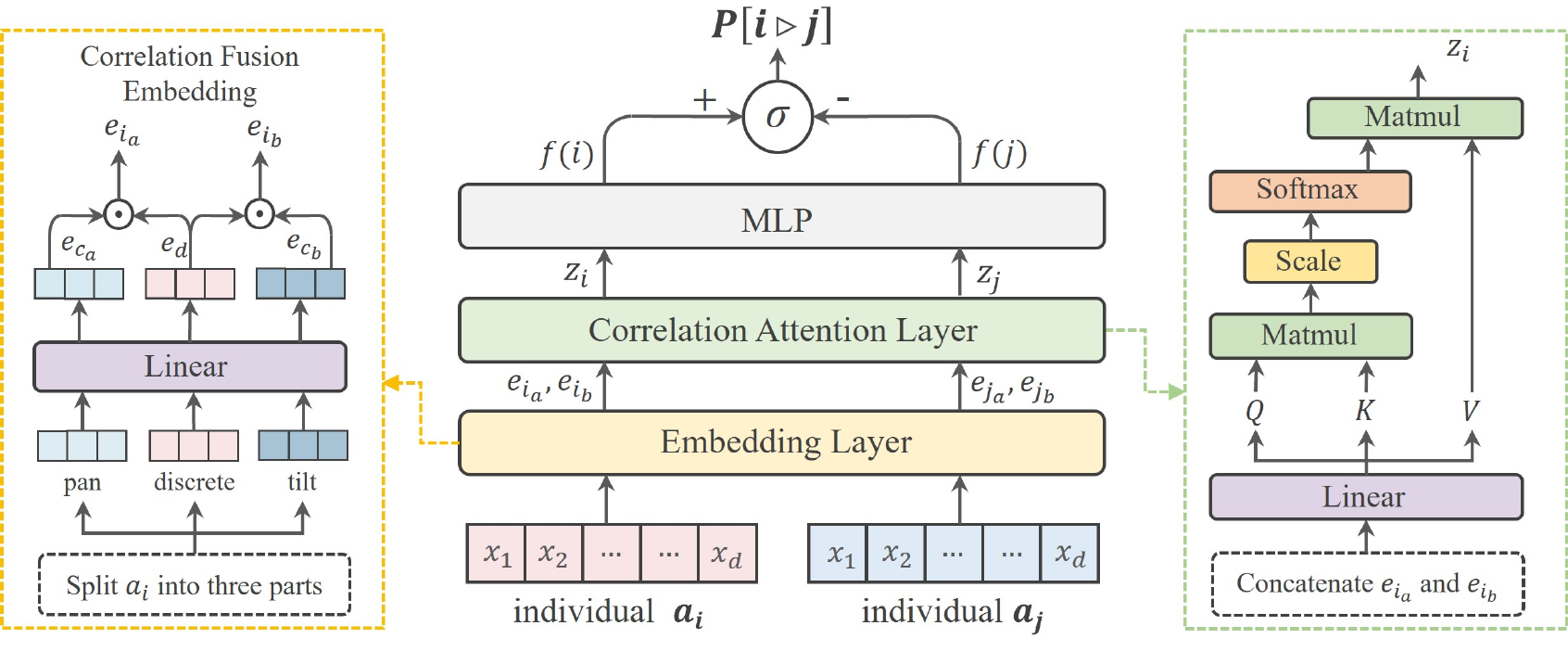}
    \caption{The network structure of global surrogate model.}
    \label{fig:global surrogate}
\end{figure}

Inspired by this framework, we build the pairwise global surrogate based on the RankNet. Given the remarkable advances in deep learning over recent years, we employ deep neural networks to model the relative preference between individuals. Fig.  \ref{fig:global surrogate} illustrates the overall structure of the global surrogate model. The model is lightweight, with the decision variables of individuals $i$ and $j$ as inputs and the output indicating which individual has better quality. Specifically, it models the ranking function through an embedding layer, a correlation attention layer, and two Multilayer Perceptron (MLP) layers.

For our mixed-variable coverage optimization, 
each solution is encoded as a 0/1 string for candidate site selection, along with the pan and tilt angular parameters. These solutions are then embedded into 5-dimensional vectors through linear projection. Given the homogeneity of many decision variables, the network shares embedding parameters to reduce the overall parameter count.

As discussed in Section \ref{sec:problem}, there is a strong correlation between discrete and continuous variables in the coverage optimization task. Therefore, we apply the multi-head attention mechanism \cite{vaswani2017attention} to capture the intricate relationships between variables effectively. For individual $i$, we use linear projections to obtain the embedding $e_{i}$, which includes the discrete part $e_{d}$ and two continuous parts $e_{c_{a}}$ and $e_{c_{b}}$. We then fuse the corresponding information with the Hadamard product. The output of this layer is subsequently delivered to the following MLPs to enhance the capacity of representing nonlinearity. All activation functions in the network are applied with ReLU, and batch normalization is implemented with PyTorch's default parameters. %Meanwhile, we do not employ dropout, following a similar approach in \cite{snoek2015scalable}.% 
The pipeline of the global surrogate can be summarized as follows.
\subsubsection{Initiate the Surrogate}
To initiate the global surrogate, we construct a training set $\mathcal{T}$ using the evaluated solutions from the archive $\mathcal{A}$. Suppose $\mathcal{A}=\{a_{1},a_{2}, ...,a_{M}\}$, where $M$ is the number of individuals in archive $\mathcal{A}$. Each training data can be represented as $\{a_{i}, a_{j}, y\},i\neq j$, where $label$ denotes the ranking between $a_{i}$ and $a_{j}$. Specifically, $y = 0$ indicates that $a_i$ is superior and vice versa. In this way, we construct the $\mathcal{T}$ with $M(M-1)$ individual pairs.

Since the generalization ability of deep networks can benefit from larger datasets  \cite{van2001art}, we introduce a novel data augmentation method to expand the datasets. According to the encoding format, the representation of individuals possesses the permutation invariance. For example, shuffling the discrete part while adjusting the continuous variables accordingly preserves the fitness ranking among individuals. Following this intuitive thought, the training dataset can be expanded at most $P_D^D$ times, effectively enhancing the network performance without significantly increasing training time. Here, we choose to expand the training set 10 times for simplicity.
% The resulting effects are discussed in the experimental section

With the training set $\mathcal{T}$, we train the global surrogate model as a binary classifier for a fixed number of $epoch=10$. The model is optimized by minimizing the cross-entropy loss function through backpropagation:
\begin{equation}
    J(W,b)=-\frac{1}{n}\sum_{i=1}^{n}{[y\cdot \log(\hat y)+(1-y)\cdot \log(1-\hat y)]},
\end{equation}
where $n$ indicates the batch size, and $\hat y$ represents the predicted output.

\subsubsection{Voting-Based Preselection Strategy}
Taking the GA variant as the global optimizer, offspring are generated iteratively. We aim to select promising solutions from the offspring for real fitness evaluation based on the trained global surrogate model. This serves as a preselection strategy within the model management of SAEAs.

In machine learning, ensemble learning reduces variance and improves model robustness by combining the predictions of multiple models. Similarly, we propose a preselection strategy based on a voting mechanism, 

\begin{figure}[!t]
    \includegraphics[width=0.92\linewidth]{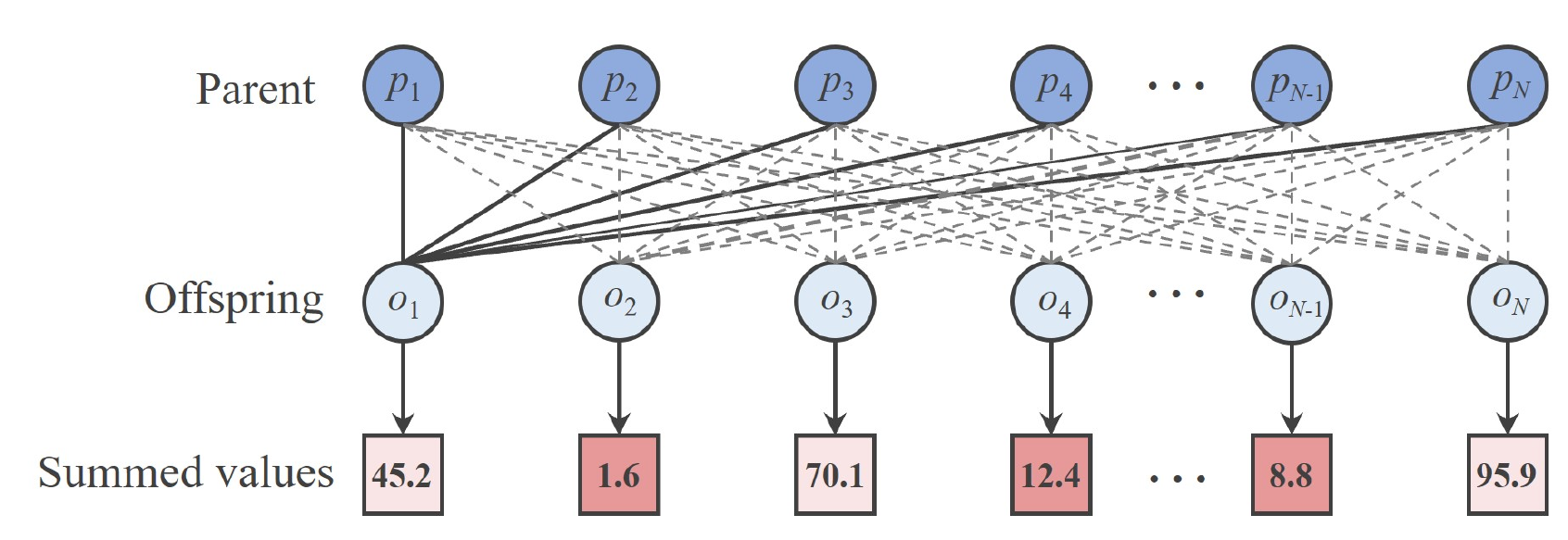}
    \caption{Diagram of voting-based preselection strategy.}
    \label{fig:preselection}
\end{figure}
For each offspring $o_{i}$, we pair it with all parent individuals to form a set of $N$ pairs. The global surrogate model then predicts for each pair of individuals $(p_{i}, o_{i})$ and sums the prediction values, denoted as $sum_{i}$. The $\tau$ offspring with the smallest summed values are selected for real fitness evaluation, with $\tau = 3$ chosen to tolerate some model error.

\subsubsection{Surrogate Update Mechanism}
As the optimization proceeds, the population evolves within the solution space, which deviates from its initial distribution. Updating the surrogate model at the proper time improves accuracy while reducing unnecessary computation. Unlike the surrogates (i.e., RBFN, Kriging) used in most SAEAs, our model is updated only when the number of newly evaluated individuals added to $\mathcal{A}$ exceeds a threshold $T$ and the global optimum is updated.

We define the set $New=\{a_{new,1}, a_{new,2},...,a_{new,T}\}$ to include the newly evaluated individuals. To mitigate catastrophic forgetting and enrich the training data, we sample 
$T$ old individuals from the archive $\mathcal{A}$  with a certain probability, forming the set $Old=\{a_{old,1}, a_{old,2},...,a_{old,T}\}$ accordingly. These old and new individuals are paired as $[a_{old, i}, a_{new, j}]$, which generates $T^2$ unseen pairs for model updating. Since the training data during updating is sparse, we also apply the data augmentation technique to enlarge the training data. Due to the archive $\mathcal{A}$ has a capacity limit, we sample only from the most recent $T_{max}$ individuals to support online learning. During the update phase, the model is trained for a fixed number of $epoch=10$.

\subsection{Surrogate-Assisted Local Estimation of Distribution}
Numerous studies have investigated that global surrogate models often fail to provide accurate approximations in local regions, particularly in high-dimensional spaces \cite{wang2019novel}. Therefore, we construct a local surrogate model based on RBFN and employ the weighted EDA as the local optimizer.

As a representative of distance-based regression models, RBFN offers powerful local approximation capabilities when paired with an appropriate kernel function. In line with approaches from other studies on mixed-variable SAEAs \cite{li2021surrogate,liu2023surrogate}, we adopt the Gaussian kernel and use the Gower distance as 
the distance metric in the kernel. For the local optimization, we adapt the discrete variable update mechanism of the weighted EDA \cite{li2021surrogate} to our fixed-size subset selection problem. 

\begin{algorithm}[!t]
   \caption{Procedure of surrogate-assisted local EDA}
   \label{alg2}  % 标签
   \setstretch{1.1}
   \begin{algorithmic}[1]%一行一个标行号
   \REQUIRE The data archive $\mathcal{A}$; \\
       \qquad Selecting number of top individuals $N_{b}$; \\ 
       \qquad Sampling number of new individuals $N_{s}$; \\
       \qquad Dimension of discrete variables $N_{d}$; \\
       \qquad Dimension of continuous variables $N_{c}$;
   \ENSURE Selected $\tau$ solutions; \\
   \STATE Select $N_{b}$ best individuals from $\mathcal{A}$ as $Pop$, and their reversed fitness value as $fit$;
   \STATE Calculate the fitness-weighted value $w=fit/{\sum_{i=1}^{N_b}{fit}}$;
   \STATE Generate statistic value for each continuous variable $j$ with  $\mu_{j}=\sum_{i=1}^{N_b}{w_i \cdot Pop_{i,j}}, \sigma_{j}=\sqrt{\frac{\sum_{i=1}^{N_b}(Pop_{i,j}-\mu_{j})^2}{N_b}}$;
   \STATE Calculate the selected probability for each discrete variable $j$ as $prob_{j}=\sum_{i=1}^{N_b}{w_{i} \cdot I(Pop_{i,j}, 1)}$;
   \STATE $X \leftarrow \emptyset$;
   \FOR{$i=1$ to $N_{s}$} 
       \STATE Sort $prob$ in descending order;
       \STATE Sample the discrete part $d_{i}$ sequentially based on $prob$ until it contains $k$ ones;
       \FOR{$j=N_{d}$ to $N_{d}+N_{c}$} 
       \STATE $c_{i, j} \leftarrow N(\mu_{j}, \sigma_{j}^2)$;
       \ENDFOR
       \STATE $X \leftarrow \{X, [d_{i}, c_{i}]\}$;
   \ENDFOR
   \STATE Train the local RBFN with $5\cdot(N_c+N_d)$ best data from $\mathcal{A}$, and select the best predicted and most uncertain solution from generated $X$;
   \end{algorithmic}
\end{algorithm}
The algorithm details are provided in Algorithm \ref{alg2}. Lines 1–4 compute the fitness weights for EDA and define the associated statistical model. In line 4, $I(a, b)$ is set to 1 if $a=b$, and 0 otherwise. Lines 6–13 describe the sampling procedure for discrete and continuous variables, which are combined to form $X$. [$\cdot$] indicates the operation of concatenation, and $N_{d}+N_{c}$ equals the total dimension $D$. In line 14, the trained RBFN is used to preselect $\tau$ individuals: the best individual is selected for exploitation, while the one with the greatest Euclidean distance from the current population, indicating the highest uncertainty, is chosen for exploration. In other words, here $\tau=2$.
\begin{figure}[!t]
\vspace{-0.4cm}
   \subfloat[Solution $a_{1}$ with $f=25.8$]{%
       \includegraphics[width=0.38\linewidth]{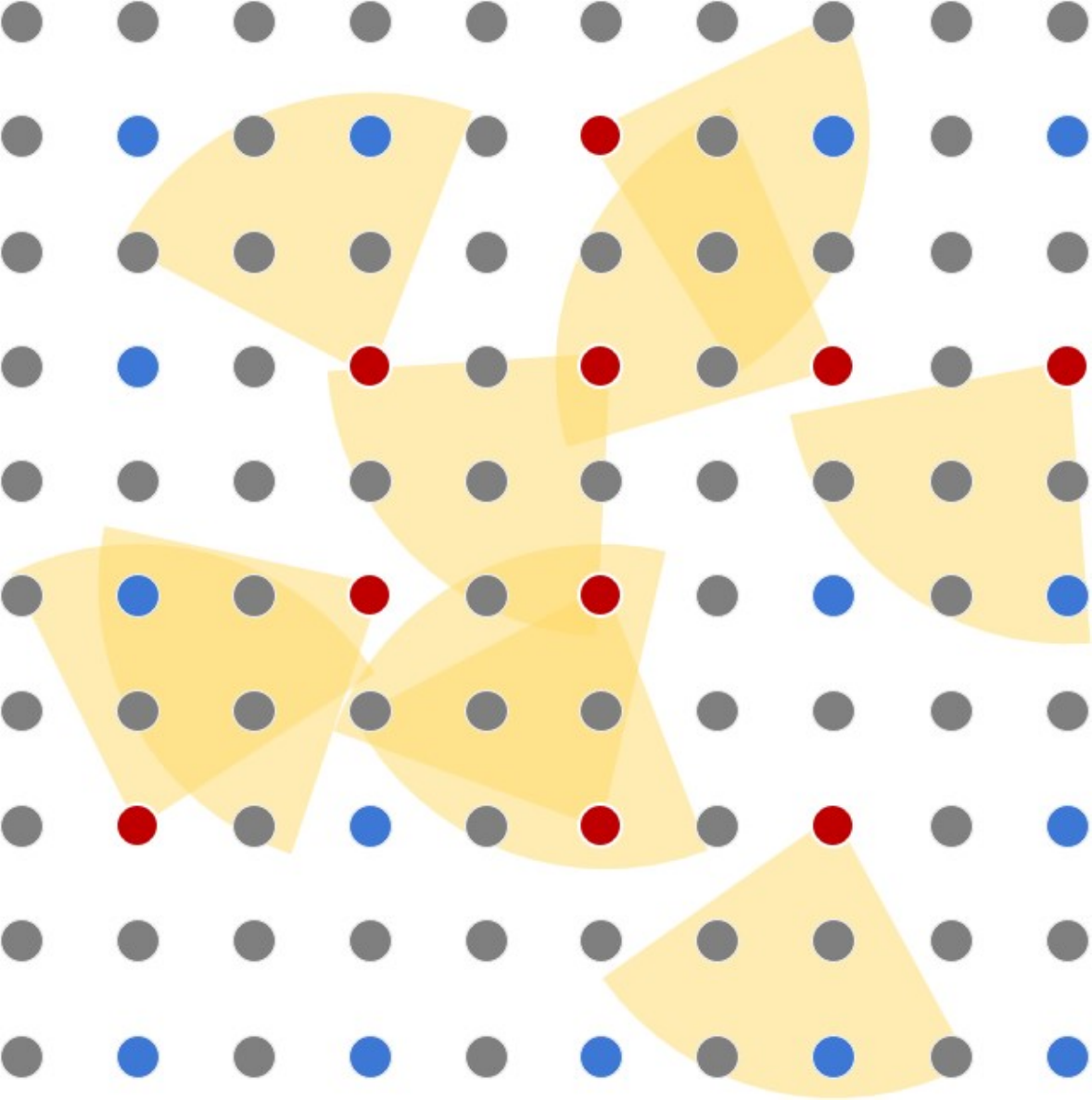}
       \label{fig:multi_modal_a}}
   \hfill
    \subfloat[Solution $a_{2}$ with $f=25.4$]{%
       \includegraphics[width=0.4\linewidth]{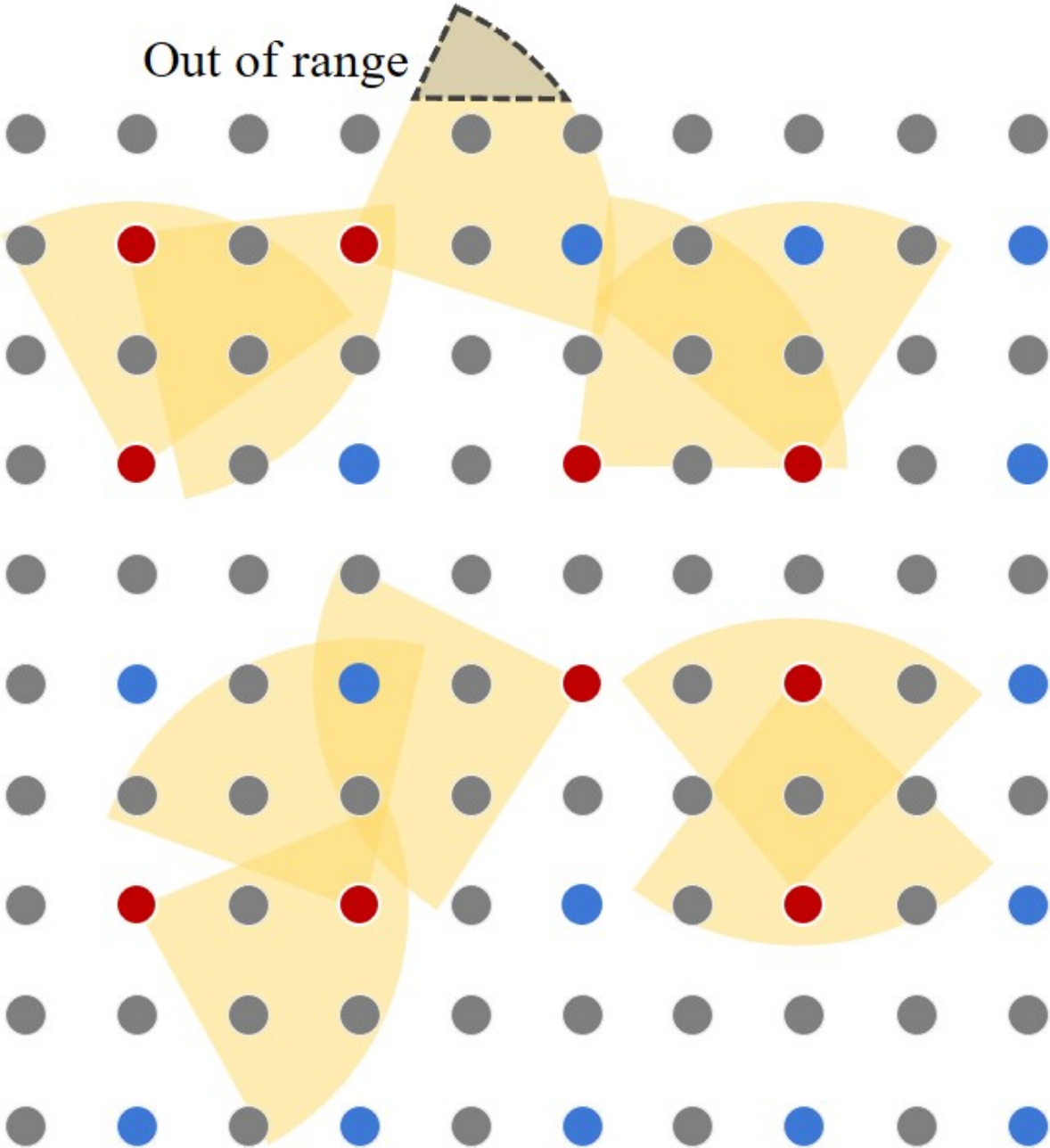}
       \label{fig:multi_modal_b}}
    \caption{Diagram of multiple plateau characteristics in our problem. Solutions with similar fitness values may differ significantly in the solution space.}
    \label{fig:multiple plateau}
\end{figure}

\vspace{-0.1cm}
\subsection{Fitness Diversity-Driven Switching Strategy}
In the RI-SHM framework, global and local optimizations play distinct roles that complement each other. Global optimization focuses on exploring the mixed-variable solution space, while local optimization aims to refine promising solutions or escape from local optima when the global optimum stagnates. To push the solution towards better boundaries, this subsection introduces a switching strategy that effectively balances global and local optimization.

Population diversity is commonly used as an indicator for adaptively switching between global and local optimization. However, this metric may not be suitable for our deployment problem. Fig. \ref{fig:multiple plateau} illustrates a top-down view showing two solutions with similar fitness values. However, their discrete components and directional orientations differ significantly in the solution space. The experimental results also reveal that the population diversity metric for MVOPs, as proposed by \cite{polakova2019differential}, remains relatively high even during evolutionary stagnation. This highlights the importance of capturing fitness diversity in coverage optimization with multiple plateaus. 

Therefore, we adopt the fitness diversity evaluation method proposed by \cite{neri2012memetic}, which is straightforward and effective.
\begin{equation}
    FD=1-|\frac{f_{avg}-f_{best}}{f_{worst}-f_{best}}|
\end{equation}
where $f_{avg},f_{worst},f_{best}$ represent the average, worst, and best fitness value among the top-ranked $N$ samples in archive $\mathcal{A}$. This approach has shown promising results in memetic algorithms \cite{neri2012memetic}.
% \cite{li2023surprisingly}

\begin{figure}[!t]
    \centering
    \includegraphics[width=1.0\linewidth]{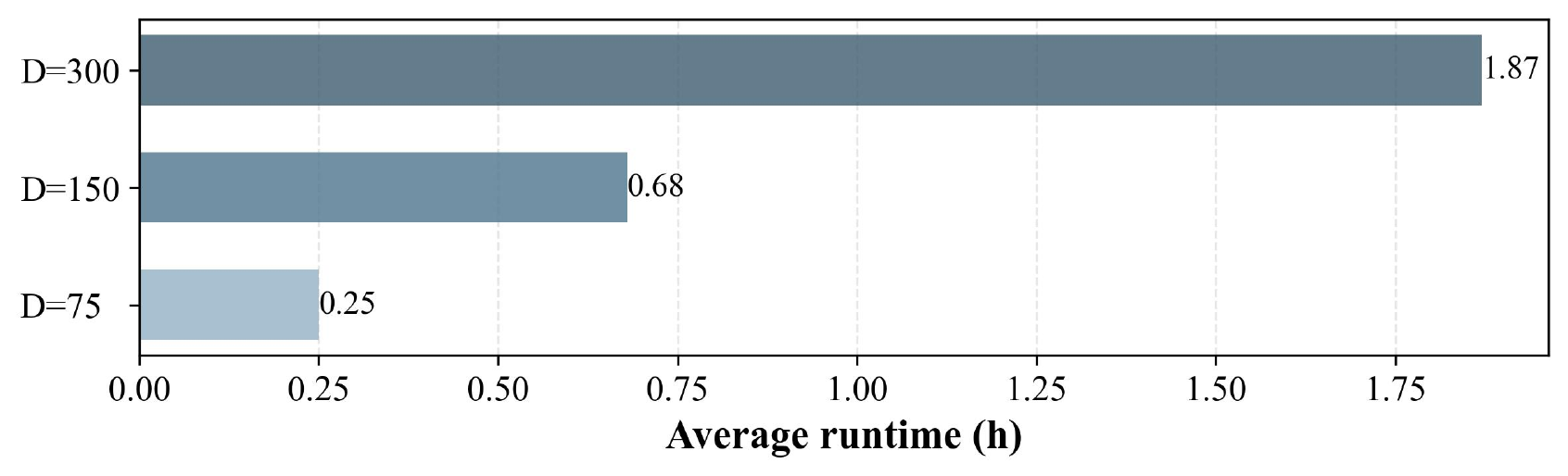}
    \caption{Running time of RI-SHM across different dimensions.}
    \label{fig:RankNet_time}
\end{figure}

Although fitness diversity may not accurately reflect the status of the population, it provides an efficient way to decide the correct moment for switching between different strategies. When the fitness diversity is low, solutions are clustered in a small region of the solution space or distributed across plateaus with similar performance. In such cases, activating an alternative search can increase the chances of detecting fresh genotypes \cite{neri2012memetic}. In this paper, we set the threshold $\delta$ at $0.2$. If the fitness diversity falls below this value, we switch the optimizer; otherwise, we maintain the current one.
\subsection{Computational Complexity}
This part analyzes the time complexity of the complete RI-SHM framework. The primary computational cost of RI-SHM comes from training and updating the RankNet-based pairwise surrogate. Since the network's parameter count is independent of problem dimensions and data size, the computational complexity per individual can be denoted as $O(epoch)$, which represents the training cost for each fitness evaluation.

Initially, we construct $M(M-1)$ data pairs from all individuals in archive $\mathcal{A}$. After applying constant-factor data augmentation, these pairs form the training set for global surrogate initialization, where $M$ is the number of initial individuals in $\mathcal{A}$. Consequently, the computational complexity for this phase is $O(M^2\cdot epoch)$.

During the iteration, the computational cost mainly arises from constructing the RBFN in the local optimization process and the online learning updates of the global surrogate model. Since the RBFN does not require an iterative training process, its time complexity is typically $O(K\cdot N_{sample})$, where $K$ stands for the number of basis functions and $N_{sample}$ is the number of samples. For simplicity, this component is ignored as $K<<N_{sample}, N_{sample}<<M^2$. The training set for the global surrogate's online learning is derived from real evaluated individuals, which are preselected within the overall framework under a limited budget of $MaxFEs$. Based on the update criterion, we assume that exactly $T$ new individuals are added to the archive $\mathcal{A}$ when the global optimum is updated, and the training set contains $T^2$ data pairs. The model can be updated up to $MaxFEs/T$ times, with each individual trained for constant epochs. Thus, the computational complexity of this component is $O(T \cdot MaxFEs \cdot epoch)$.

Based on the above analysis, the overall computational complexity of RI-SHM is $O(M^2\cdot epoch+T \cdot MaxFEs \cdot epoch)$. In our experiments, $T<<M$ and the magnitude of $T \cdot MaxFEs$ is much smaller than $M^2$. Therefore, the final computational complexity of RI-SHM is $O(M^2\cdot epoch)$. It is evident that, given a fixed training cost of $O(epoch)$ per individual, RI-SHM's computational complexity is related to the square of the number of initialized individuals $M^2$, where $M=2D$. This indicates that the complexity scales with the problem's dimensionality, which can be observed from Fig. \ref{fig:RankNet_time}.

\section{Experimental results}
\label{sec: experiment}
In this section, we outline the basic experimental settings. Next, we evaluate and validate the performance of RI-SHM across small to large-scale deployment test instances. Finally, we conduct ablation studies to verify the effectiveness of RI-SHM's core components.

\subsection{Experimental Design}
\subsubsection{Test Instances} 
\begin{figure}[!t]
    \centering
   \subfloat[Set up]{%
       \includegraphics[width=0.52\linewidth]{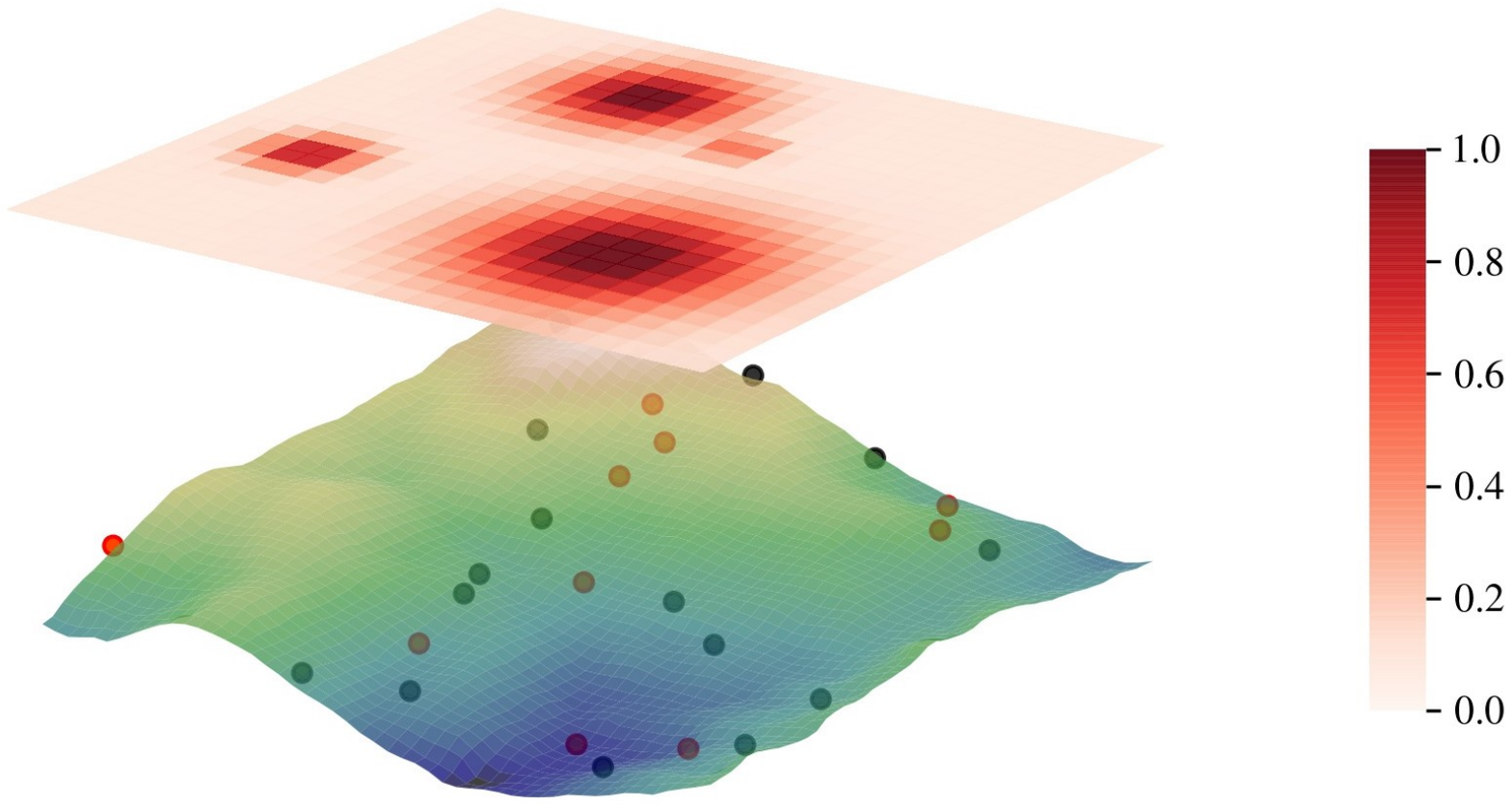}
       \label{fig:coverage_a}}
    \hfill
    \subfloat[Coverage probability]{%
       \includegraphics[width=0.42\linewidth]{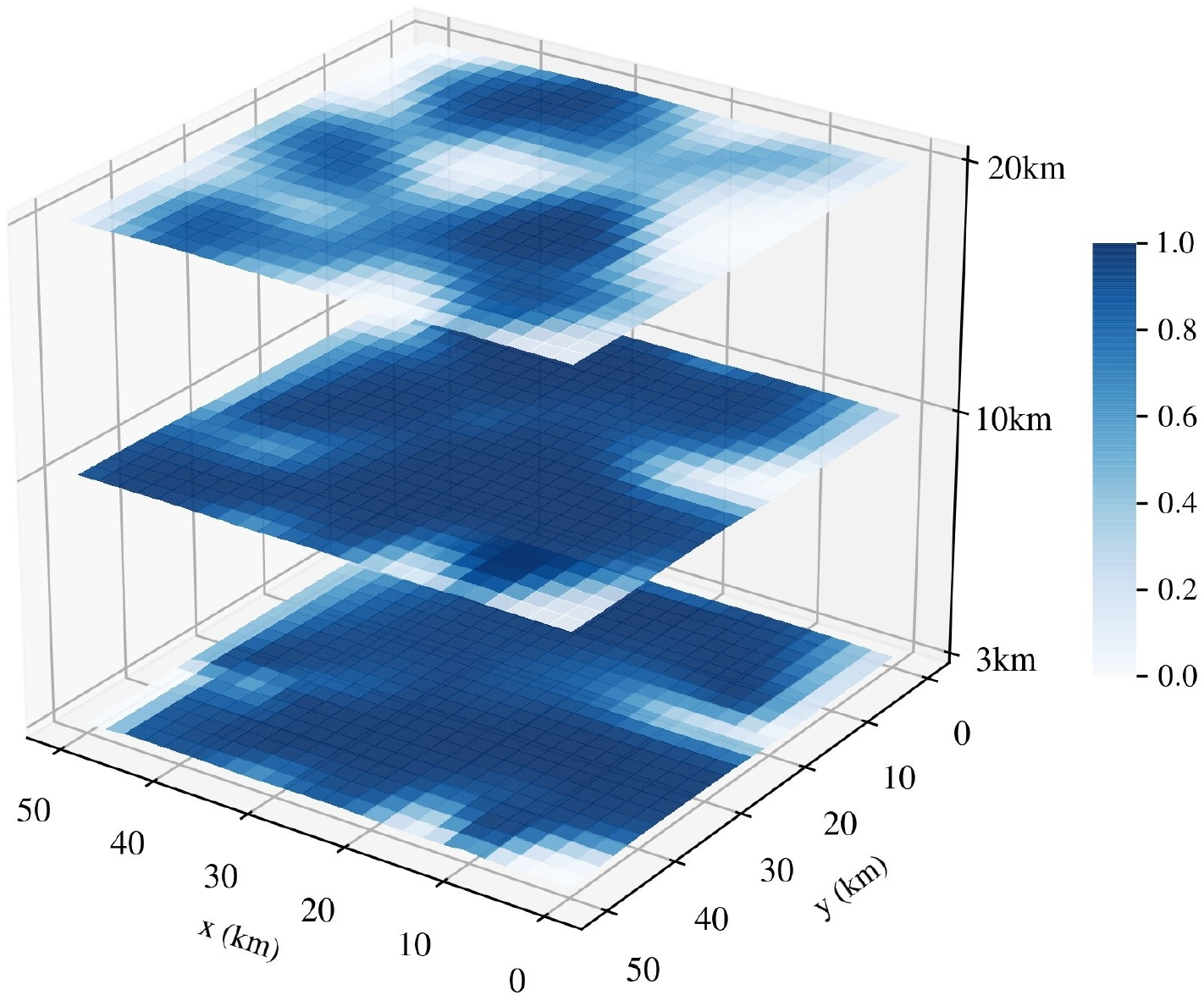}
       \label{fig:coverage_b}}
    \caption{The experimental environment alongside a visualization of the coverage achieved by RI-SHM, where targets closer to the critical areas receive higher weights.}
    \label{fig:example}
\end{figure}

In this study, we focus on the coverage optimization for target coverage in 3-D mountainous environments. To simulate a realistic application scenario, we use open-source DEM data\footnote{https://portal.opentopography.org/datasets} covering a 50km$ \times $50km region in Beijing, China, as the deployment area. The aerial space is divided into three altitude levels at 3km, 10km, and 20km, consistent with the parameters used in \cite{lian2012three, zhang2023surrogate, wu2024mixed}. The target set $Q$ is uniformly sampled from these layers with varying levels of precision. Each target $q \in Q$ is assigned with a weight $\omega_{q}$, where targets closer to the critical areas receive higher weights. Fig. \ref{fig:example} illustrates the experimental setup and demonstrates the coverage performance achieved by RI-SHM.

Given the limited number of sensors, we assume that the deployment resources are restricted to 10 sensors, i.e., $k=10$, and each sensor has identical parameters: $\beta_{d}=1, \beta_{p}=0.15, \beta_{t}=0.15, t_d=25, t_p=40$, and $t_t=40$.

To validate the effectiveness of the proposed RI-SHM, we consider test scenarios at small, medium, and large scales, with parameters detailed in Table \ref{tab:setting}. Each scale includes five instances, differing in their candidate site sets. These candidate sets are randomly sampled from the deployment area.

\begin{table}[!t]
\vspace{-0.15cm}
\caption{Parameter settings of different scales}
\label{tab:setting}
\centering
\scriptsize
\begin{tabular}{c|cccc}
\toprule
\textbf{Scale}  & \begin{tabular}[c]{@{}c@{}}\textbf{Dimension}\\ $D$\end{tabular} & \begin{tabular}[c]{@{}c@{}}\textbf{Candidate sites}\\ $|Z|$\end{tabular} & \begin{tabular}[c]{@{}c@{}}\textbf{Deployed sensors}\\ $k$\end{tabular} & \begin{tabular}[c]{@{}c@{}}\textbf{Target}\\ $|Q|$\end{tabular} \\ \midrule
Small  & 75         & 25                                                            & 10                                                           & 300                                                  \\
Medium & 150         & 50                                                            & 10                                                           & 867                                                  \\
Large  & 300         & 100                                                           & 10                                                           & 1875                                                 \\ \bottomrule
\end{tabular}
\end{table}

\begin{table}[!t]
\vspace{-0.15cm}
\caption{Parameter settings for RI-SHM}
\label{tab:para}
\centering
\footnotesize
\begin{tabular}{c|l|c}
\toprule
\textbf{Module}            & \multicolumn{1}{c|}{\textbf{Paramters}}   & \textbf{Value} \\ \midrule
\multirow{5}{*}{Evolution} & Crossover Probability $P_c$           & 1.0            \\
                           &Mutation Probability of  $P_m$            & 0.1            \\
                           &Fitness Diversity Threshold  $\delta$            & 0.2            \\
                           & Number of excellent individuals in EDA $N_{b}$ & $0.45N$          \\
                           & Number of sampled individuals in EDA $N_{s}$   & $2N$             \\ \midrule
\multirow{6}{*}{Surrogate} & Embedding size $embed$                      & 5              \\
                           & Feedforward dimension $d$                  & 512            \\
                           & Number of attention heads $n_{head}$           & 8              \\
                           & Batch size $B$                              & 512            \\
                           & Learning rate $l$                           & 0.001          \\
                           & Epochs for training DNN $epoch $    & 10             \\
\multicolumn{1}{l|}{}      & Maximum size for updating $T_{max}$         & 1000           \\
\multicolumn{1}{l|}{}      & Minimum number of new individuals $T$       & 10             \\ \bottomrule
\end{tabular}
\end{table}

Due to the inherent combinatorial property, the subset selection decision becomes increasingly complex as the number of candidate locations grows. Similarly, higher sampling precision and a more significant number of targets further complicate the challenge of maximizing coverage. 

% \usepackage{multirow}
\begin{table*}[!t]
\vspace{-0.12cm}
\setlength{\tabcolsep}{1pt}
\renewcommand{\arraystretch}{1.2}
\centering
\scriptsize
\caption{Statistical results of fitness values under different scales, presented as Avg(Std), obtained by RI-SHM and four comparative algorithms. The best-performing results are highlighted.}
\label{compare:small}
\resizebox{0.95\textwidth}{!}{
\begin{threeparttable}
\begin{tabular}{c|cc|ccc|ccc|ccc|ccc|ccc}
\toprule
Scale&\multicolumn{1}{c|}{Inst.$^\diamond$} & D                   && RI-SHM                      &&& MixedEGO$^{*}$           &&& SHEDA              &&& SHEALED            &&& DEDSO        &      \\ \midrule
\multirow{5}{*}{\rotatebox[origin=c]{90}{Small}} & \multicolumn{1}{c|}{1}        & \multirow{5}{*}{75} && \cellcolor{gray!20}\textbf{1.55E+01(1.67E+00)} &&& 2.76E+01(5.26E+00) &&& 4.05E+01(3.46E+00) &&& 3.35E+01(2.72E+00) &&& 4.68E+01(4.32E+00) \\
&\multicolumn{1}{c|}{2}        &                     && \cellcolor{gray!20}\textbf{1.34E+01(1.52E+00)} &&& 2.35E+01(2.63E+00) &&& 4.25E+01(4.11E+00) &&& 3.27E+01(3.58E+00) &&& 4.33E+01(3.98E+00) \\
&\multicolumn{1}{c|}{3}        &                     && \cellcolor{gray!20}\textbf{1.44E+01(1.74E+00)} &&& 2.12E+01(3.36E+00) &&& 3.95E+01(2.94E+00) &&& 3.46E+01(2.48E+00) &&& 4.55E+01(4.60E+00) \\
&\multicolumn{1}{c|}{4}        &                     && \cellcolor{gray!20}\textbf{1.52E+01(2.32E+00)} &&& 2.45E+01(6.01E+00) &&& 4.34E+01(5.40E+00) &&& 3.53E+01(3.46E+00) &&& 4.78E+01(4.15E+00) \\
&\multicolumn{1}{c|}{5}        &                     && \cellcolor{gray!20}\textbf{1.39E+01(1.01E+00)} &&& 2.76E+01(4.54E+00) &&& 4.28E+01(9.53E+00) &&& 3.43E+01(3.77E+00) &&& 4.89E+01(3.99E+00) \\ \midrule
%\multicolumn{3}{c|}{$+/\approx/-$}                          && -                           &&& 5 / 0 / 0                &&& 5 / 0 / 0               &&& 5 / 0 / 0               &&& 5 / 0 / 0               \\ \midrule
\multirow{5}{*}{\rotatebox[origin=c]{90}{Median}}&\multicolumn{1}{c|}{1}        & \multirow{5}{*}{150} && \cellcolor{gray!20}\textbf{3.84E+01(2.45E+00)} &&& 5.01E+01(5.61E+00) &&& 1.21E+02(2.95E+01) &&& 1.01E+02(1.49E+01) &&& 1.39E+02(1.07E+01) \\
&\multicolumn{1}{c|}{2}        &                      && \cellcolor{gray!20}\textbf{3.91E+01(3.14E+00)} &&& 6.08E+01(1.07E+01) &&& 8.88E+01(8.54E+00) &&& 9.83E+01(1.24E+01) &&& 1.43E+02(7.31E+00) \\
&\multicolumn{1}{c|}{3}        &                      && \cellcolor{gray!20}\textbf{3.77E+01(1.30E+00)} &&& 6.17E+01(1.26E+01) &&& 1.29E+02(1.56E+01) &&& 9.89E+01(1.46E+01) &&& 1.39E+02(6.07E+00) \\
&\multicolumn{1}{c|}{4}        &                      && \cellcolor{gray!20}\textbf{3.83E+01(3.32E+00)} &&& 6.02E+01(1.38E+01) &&& 8.38E+01(2.36E+01) &&& 9.21E+01(6.42E+00) &&& 1.38E+02(1.30E+01) \\
&\multicolumn{1}{c|}{5}        &                      && \cellcolor{gray!20}\textbf{3.95E+01(1.81E+00)} &&& 6.57E+01(1.11E+01) &&& 1.27E+02(2.46E+01) &&& 1.05E+02(1.04E+01) &&& 1.44E+02(7.77E+00) \\ \midrule
%\multicolumn{3}{c|}{$+/\approx/-$}                           && -                           &&& 5 / 0 / 0      &&& 5 / 0 / 0                &&& 5 / 0 / 0              &&& 5 / 0 / 0               \\ \midrule
\multirow{5}{*}{\rotatebox[origin=c]{90}{Large}}&\multicolumn{1}{c|}{1}        & \multirow{5}{*}{300} && \cellcolor{gray!20}\textbf{7.86E+01(3.30E+00)} &&& \multirow{5}{*}{-} &&& 1.64E+02(3.34E+01) &&& 2.07E+02(2.35E+01) &&& 2.52E+02(1.09E+01) \\
&\multicolumn{1}{c|}{2}        &                      && \cellcolor{gray!20}\textbf{7.71E+01(5.52E+00)} &&&  &&& 1.43E+02(2.52E+01) &&& 2.25E+02(2.05E+01) &&& 2.56E+02(1.88E+00) \\
&\multicolumn{1}{c|}{3}        &                      && \cellcolor{gray!20}\textbf{7.92E+01(3.87E+00)} &&&  &&& 1.78E+02(2.87E+01) &&& 2.11E+02(1.26E+01) &&& 2.51E+02(7.69E+00) \\
&\multicolumn{1}{c|}{4}        &                      && \cellcolor{gray!20}\textbf{7.79E+01(5.00E+00)} &&&  &&& 1.75E+02(1.88E+01) &&& 2.18E+02(2.14E+01) &&& 2.53E+02(6.60E+00) \\
&\multicolumn{1}{c|}{5}        &                      && \cellcolor{gray!20}\textbf{8.14E+01(5.00E+00)} &&&  &&& 1.87E+02(3.10E+01) &&& 2.28E+02(2.74E+01) &&& 2.59E+02(9.87E+00) \\ \midrule
\multicolumn{3}{c|}{$+/\approx/-$}                           && -                           &&& 14 / 0 / 0    &&& 14 / 0 / 0              &&& 14 / 0 / 0              &&& 14 / 0 / 0              \\ \bottomrule
\end{tabular}
\begin{tablenotes}    
    \item \scriptsize $^\diamond$ Inst. is the shorten for Instance.
    \item \scriptsize $^{*}$ Notably, MixedEGO is excluded from large-scale comparisons as it fails to solve these instances within 7 days due to high computational costs.
\end{tablenotes} 
\end{threeparttable}}
\end{table*}

\subsubsection{Algorithms in Comparison} We compare the proposed RI-SHM with the following regression-based algorithms, as no classification- or pairwise-based SAEAs have been developed for MVOPs. The effectiveness of the pairwise-based model is discussed in Section \ref{compare:impact}.
\begin{enumerate}[a)]
    \item \textit{MixedEGO} \cite{saves2024smt}: A robust algorithm adopts the Bayesian optimization framework, featuring the Kriging surrogate with a kernel specifically tailored for MVOPs\footnote{Implemented using the Python package SMT.}. The infill criterion follows the Expected Improvement (EI) maximization rule, consistent with the classical Efficient Global Optimization (EGO) \cite{jones1998efficient}. 
    \item \textit{SHEDA} \cite{li2021surrogate}: An efficient algorithm belongs to EC-based optimization, frequently used as a comparative algorithm in EMVOPs. It combines EDA with canonical Kriging to solve NAS.
    \item \textit{SHEALED} \cite{liu2023surrogate}: A novel algorithm\footnote{https://github.com/HandingWangXDGroup/SHEALED} adopts the global-local SAEA framework and employs RBFN as its surrogate model. In addition to global and local optimizers, it incorporates a local continuous search module to refine continuous decision variables in the late stage.
    \item \textit{DEDSO} \cite{xie2023dual}: The latest algorithm\footnote{https://github.com/ForrestXie9/DEDEO} that presents an SAEA specifically for MVOPs, using $ACO_{mv}$ \cite{liao2013ant} as the base optimizer and RBFN as the primary surrogate. It exhibits strong performance in problems with dimensions up to 100.
\end{enumerate}

All algorithms are implemented in Python and executed 10 times independently for each instance on a server with Intel(R) Xeon(R) Gold 6258R CPU @ 2.70GHz and NVIDIA GeForce RTX 4090 GPU. For RI-SHM, the deep learning model is developed using PyTorch. 
\subsubsection{Parameter Settings} In all algorithms compared, $2D$ samples are generated to initialize the data archive $\mathcal{A}$, from which the best $N$ individuals form the initial population. Here, $N$ is set to 100. The $MaxFEs$ is limited to 2000, only one-tenth of the original setting in \cite{wu2024mixed} and aligns with the setting in \cite{xiang2019clustering}.

To ensure fairness,  the parameters for MixedEGO \cite{saves2024smt}, SHEDA \cite{li2021surrogate}, SHEALED \cite{liu2023surrogate}, and DEDSO \cite{xie2023dual} are set according to the recommendations in their original papers. For RI-SHM, other parameters are shown in Table \ref{tab:para}. Note that the parameters remain unchanged across all problem scales for experimental consistency.

\begin{figure*}[!t]
\centering
    % 第一行子图
    \subfloat{%
       \includegraphics[width=0.195\textwidth]{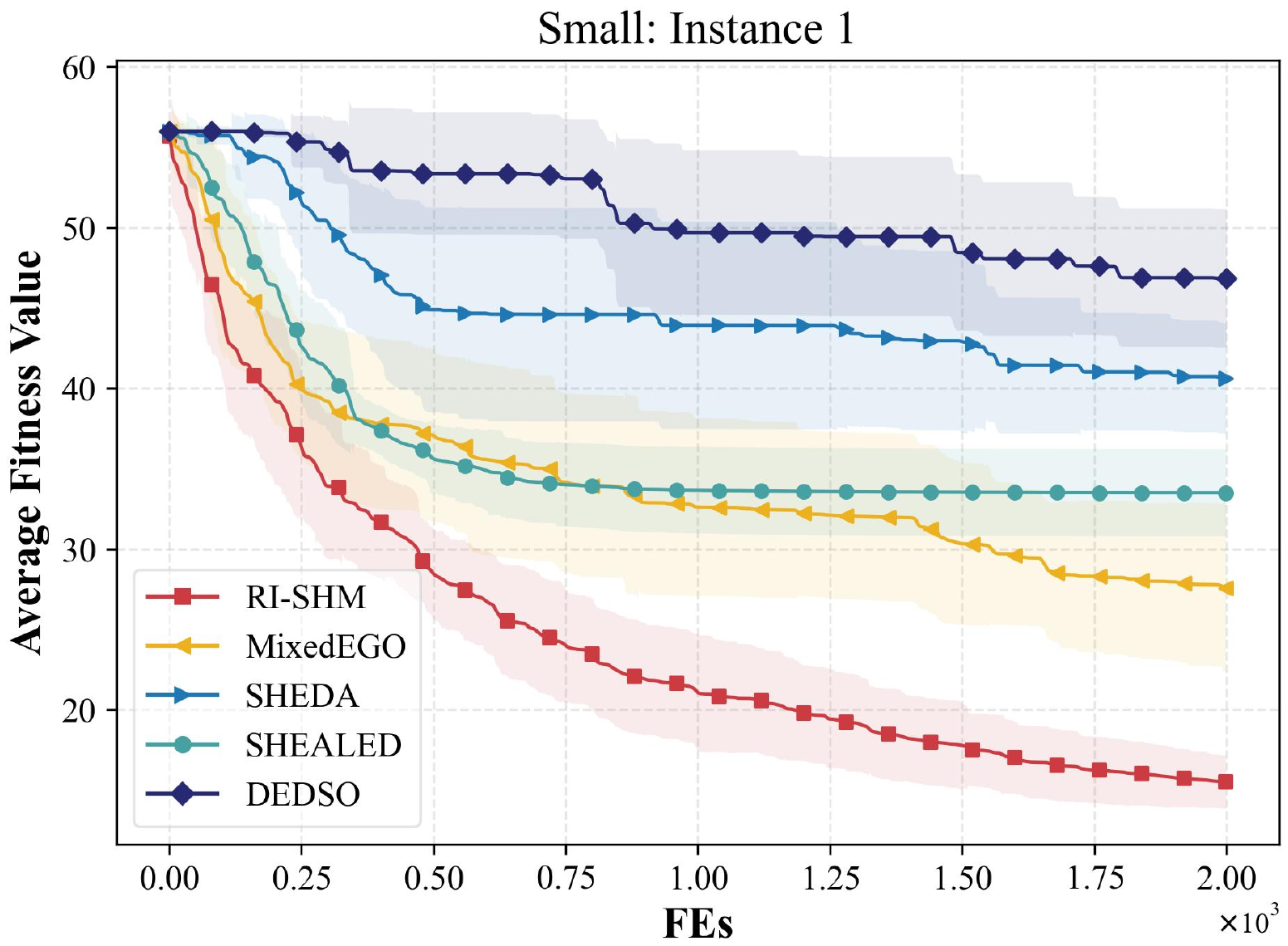}
       \label{fig:small_01}}
    \subfloat{%
       \includegraphics[width=0.195\textwidth]{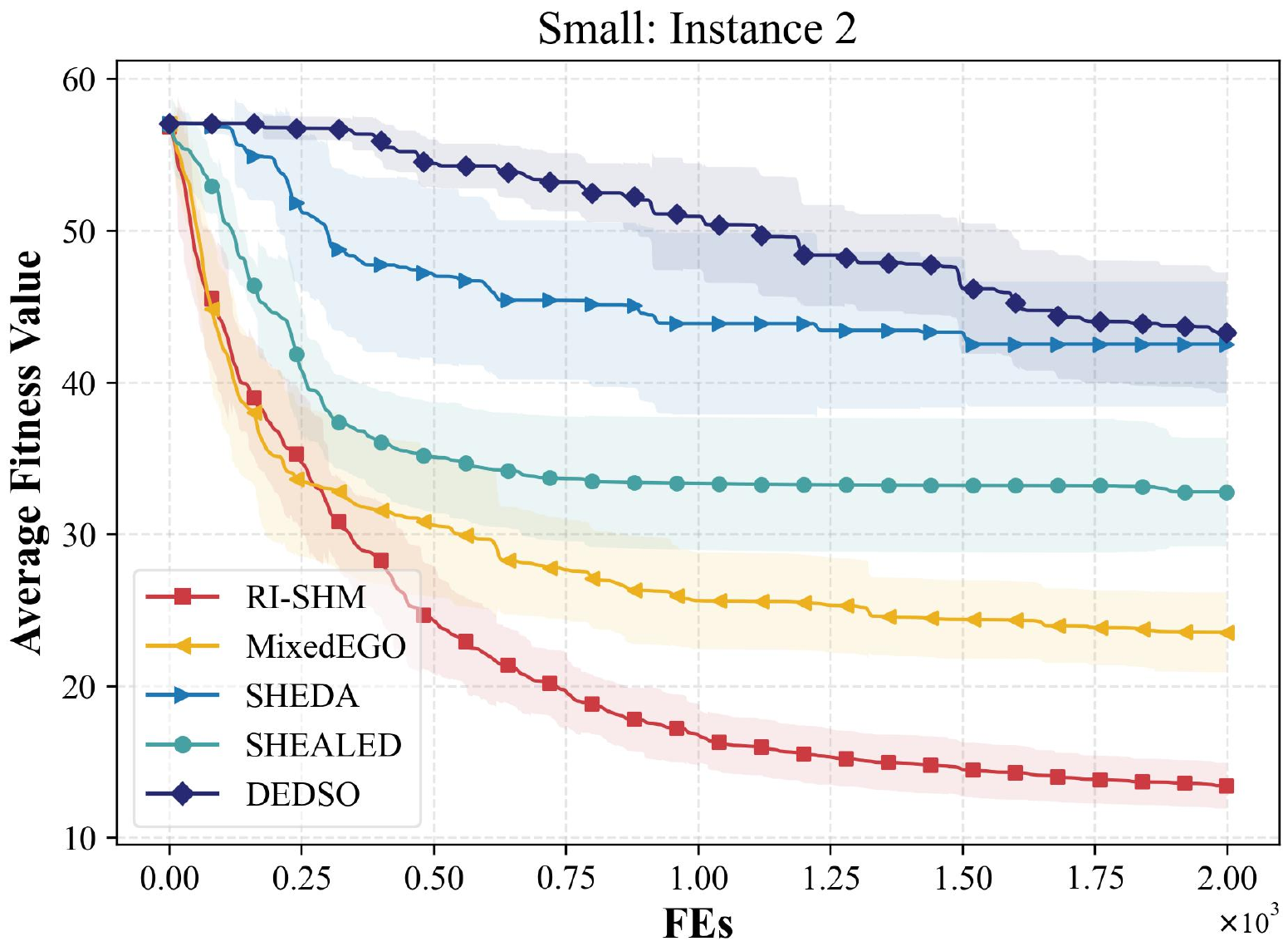}
       \label{fig:small_02}}
    \subfloat{%
       \includegraphics[width=0.195\textwidth]{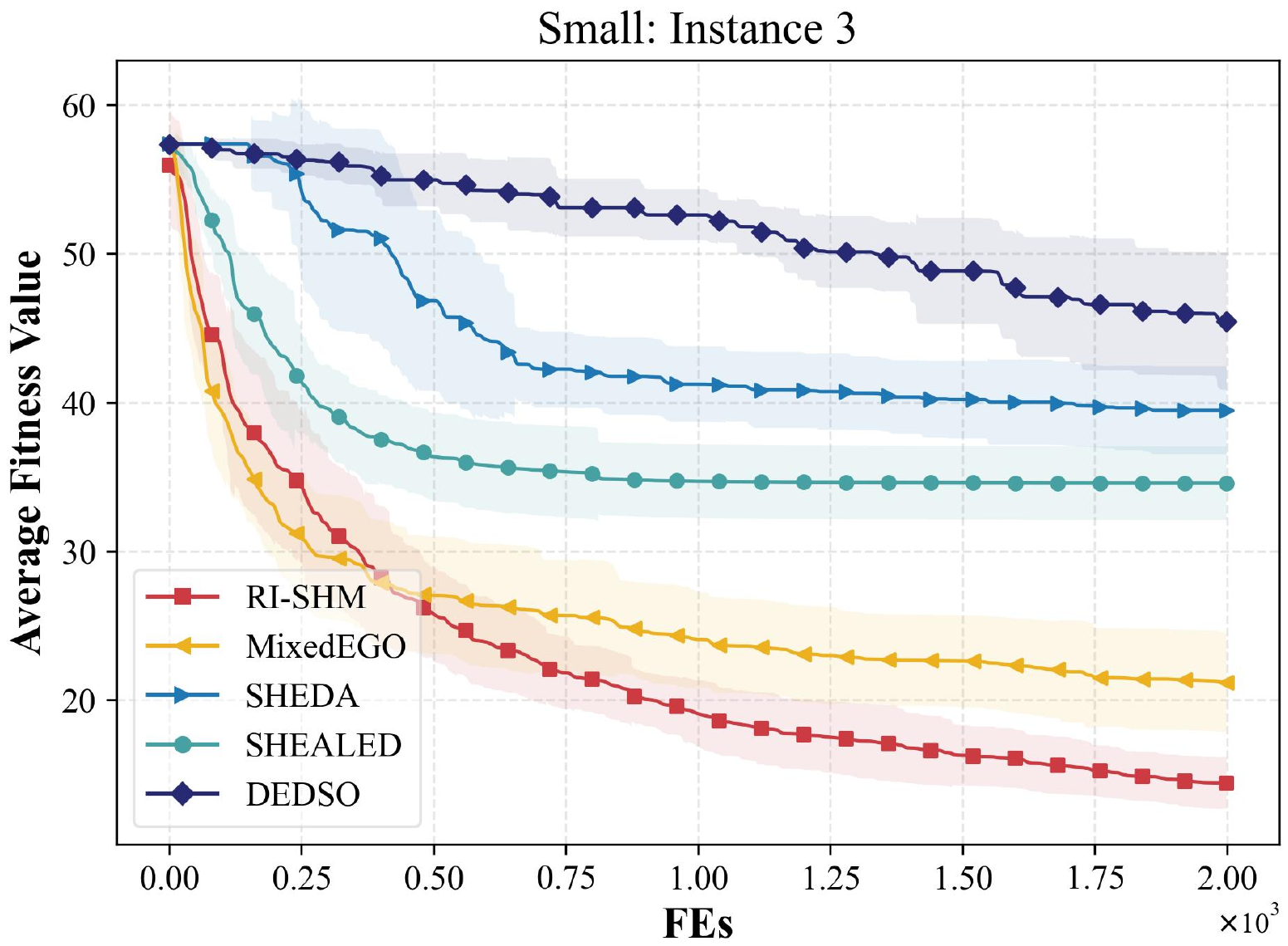}
       \label{fig:small_03}}
    \subfloat{%
       \includegraphics[width=0.195\textwidth]{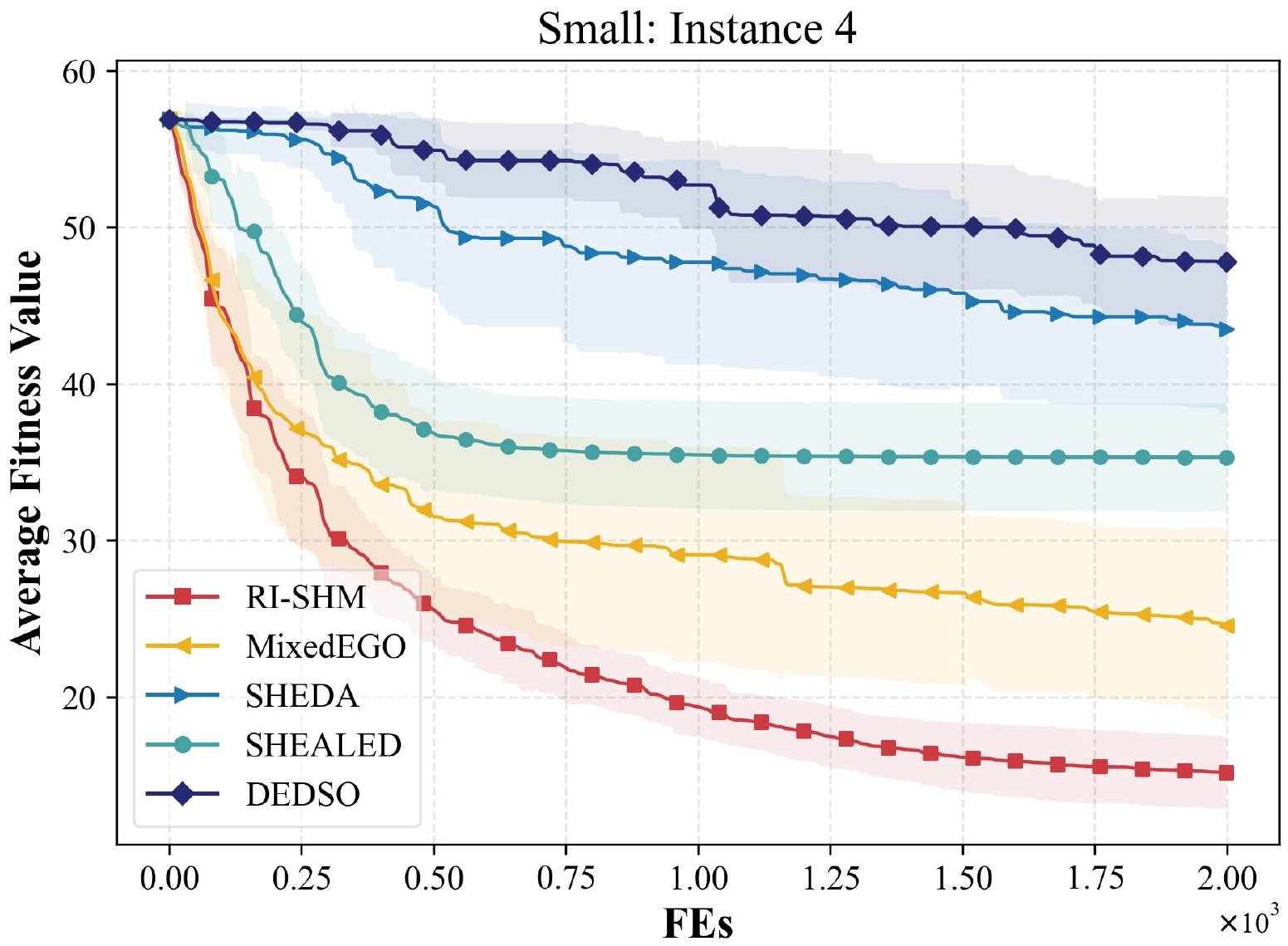}
       \label{fig:small_04}}
    \subfloat{%
       \includegraphics[width=0.195\textwidth]{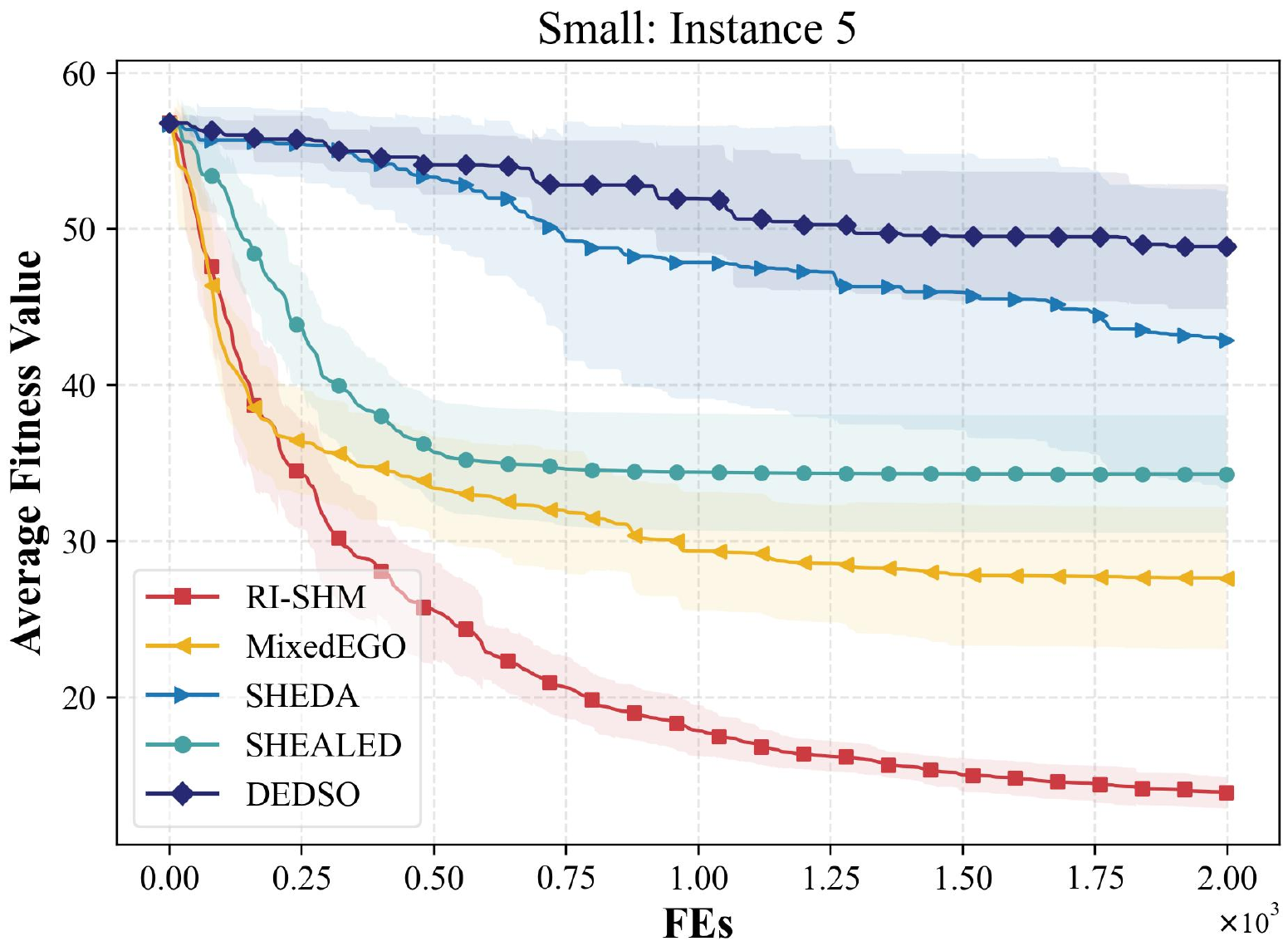}
       \label{fig:small_05}}
    \\
    % 第一行子图
    \subfloat{%
       \includegraphics[width=0.195\textwidth]{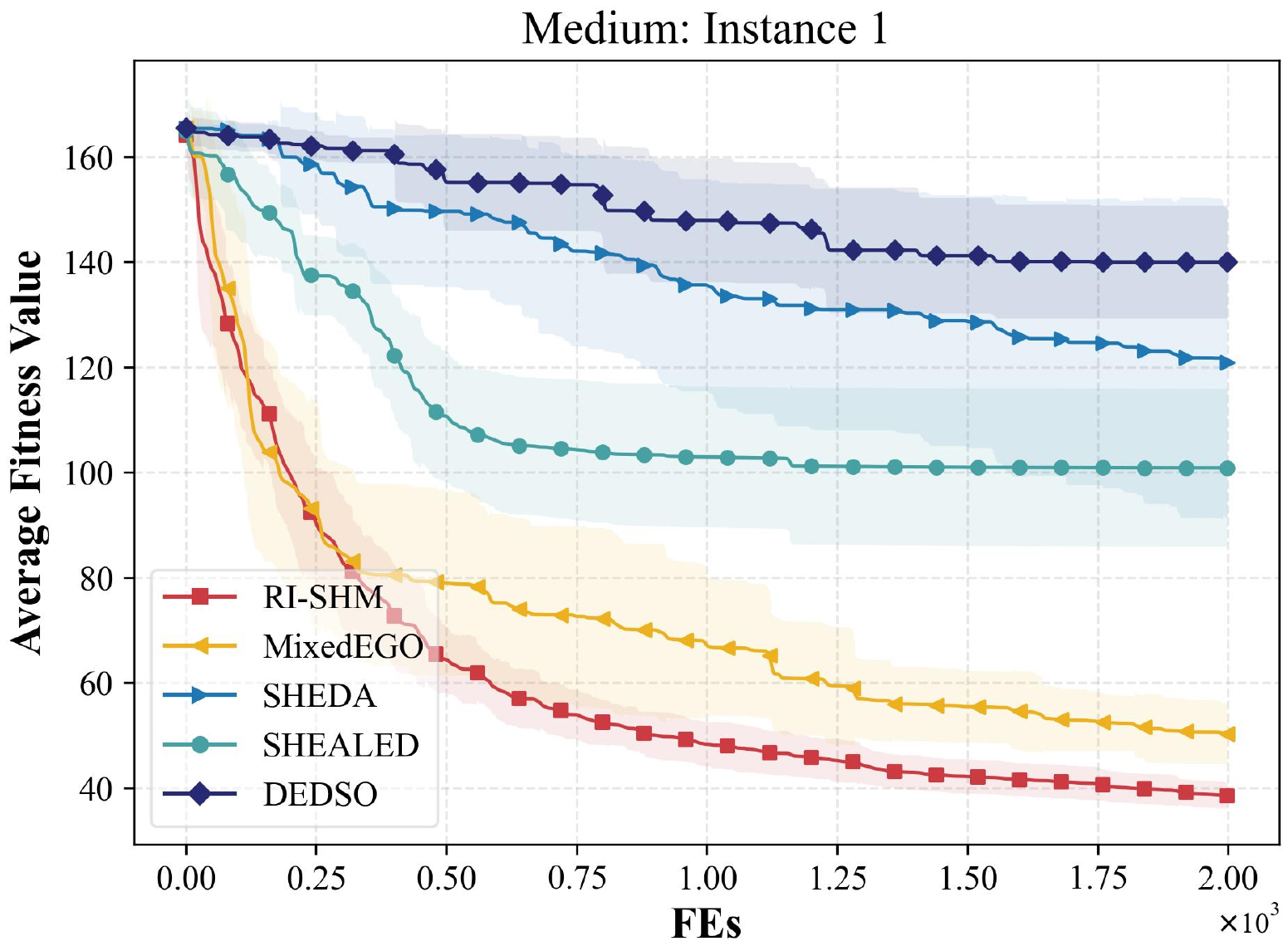}
       \label{fig:medium_01}}
    \subfloat{%
       \includegraphics[width=0.195\textwidth]{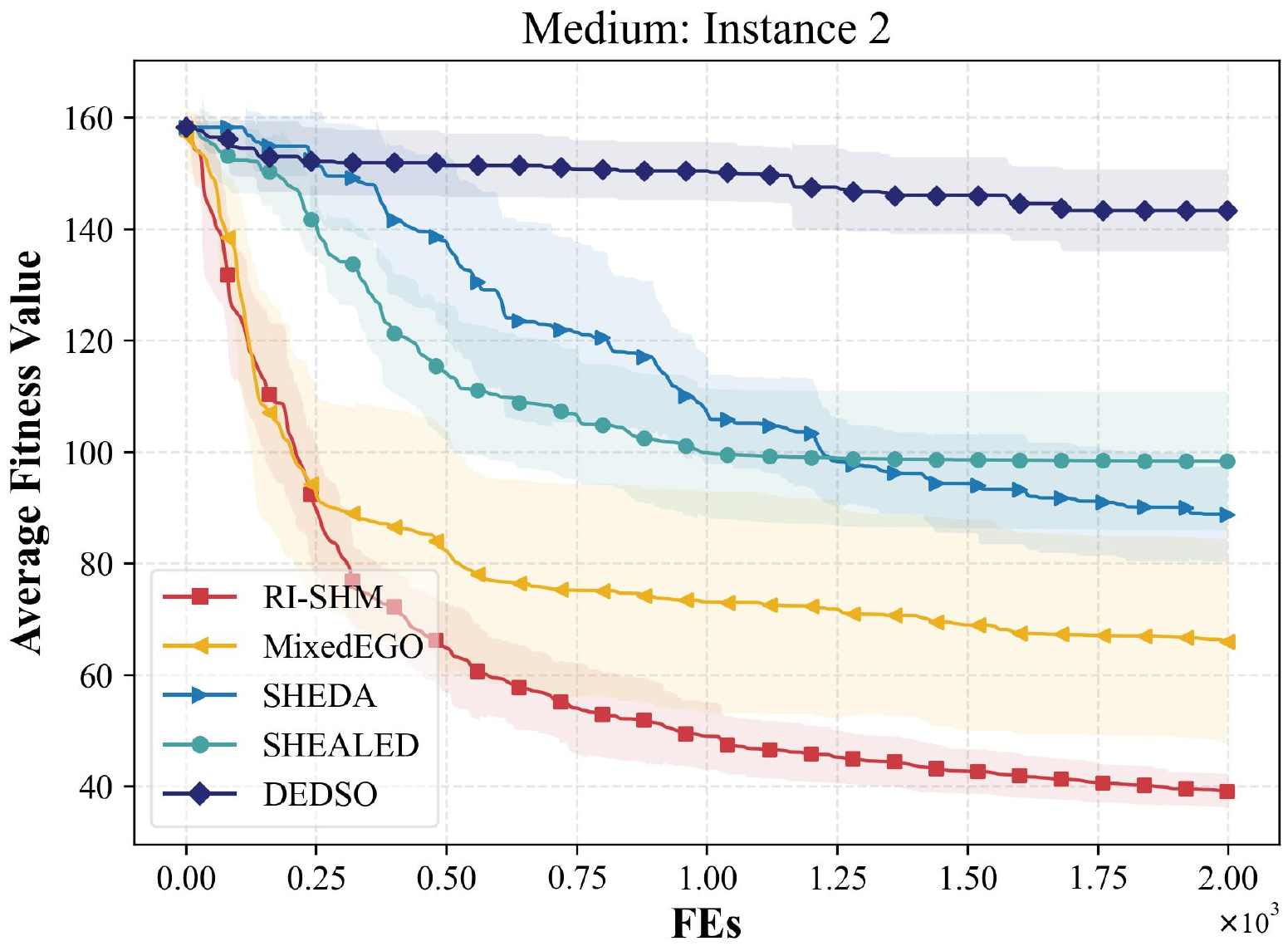}
       \label{fig:medium_02}}
    \subfloat{%
       \includegraphics[width=0.195\textwidth]{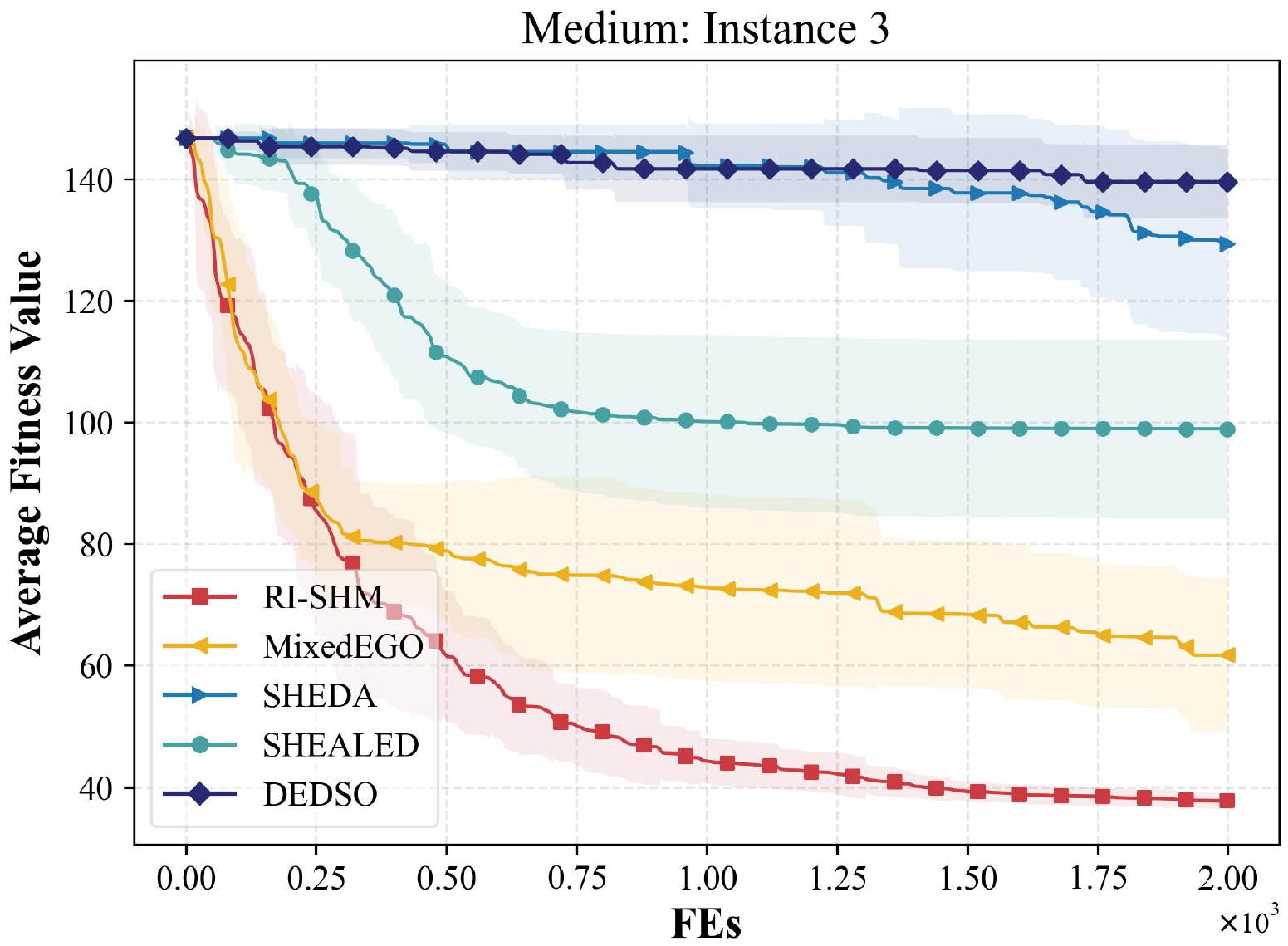}
       \label{fig:medium_03}}
    \subfloat{%
       \includegraphics[width=0.195\textwidth]{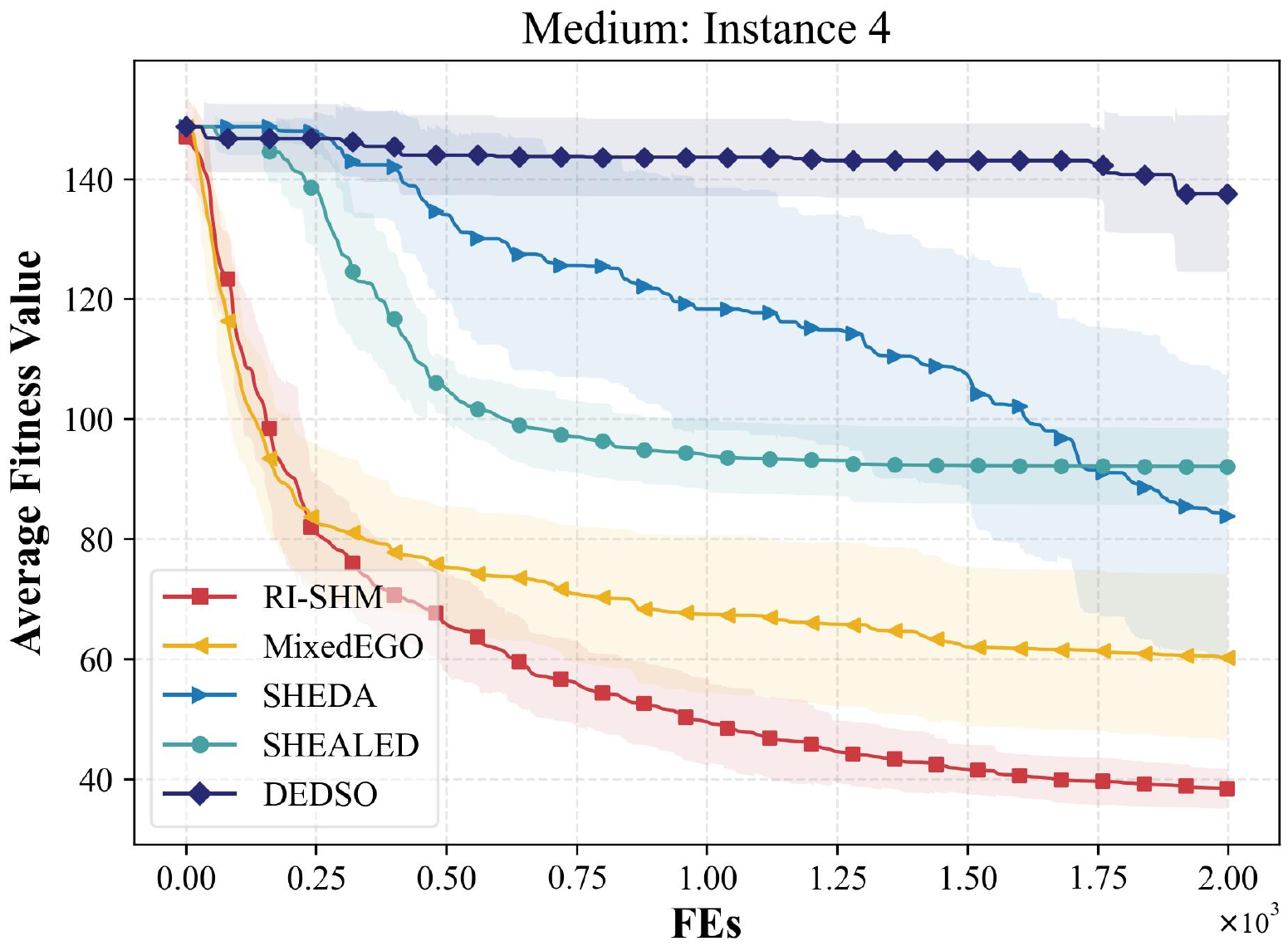}
       \label{fig:medium_04}}
    \subfloat{%
       \includegraphics[width=0.195\textwidth]{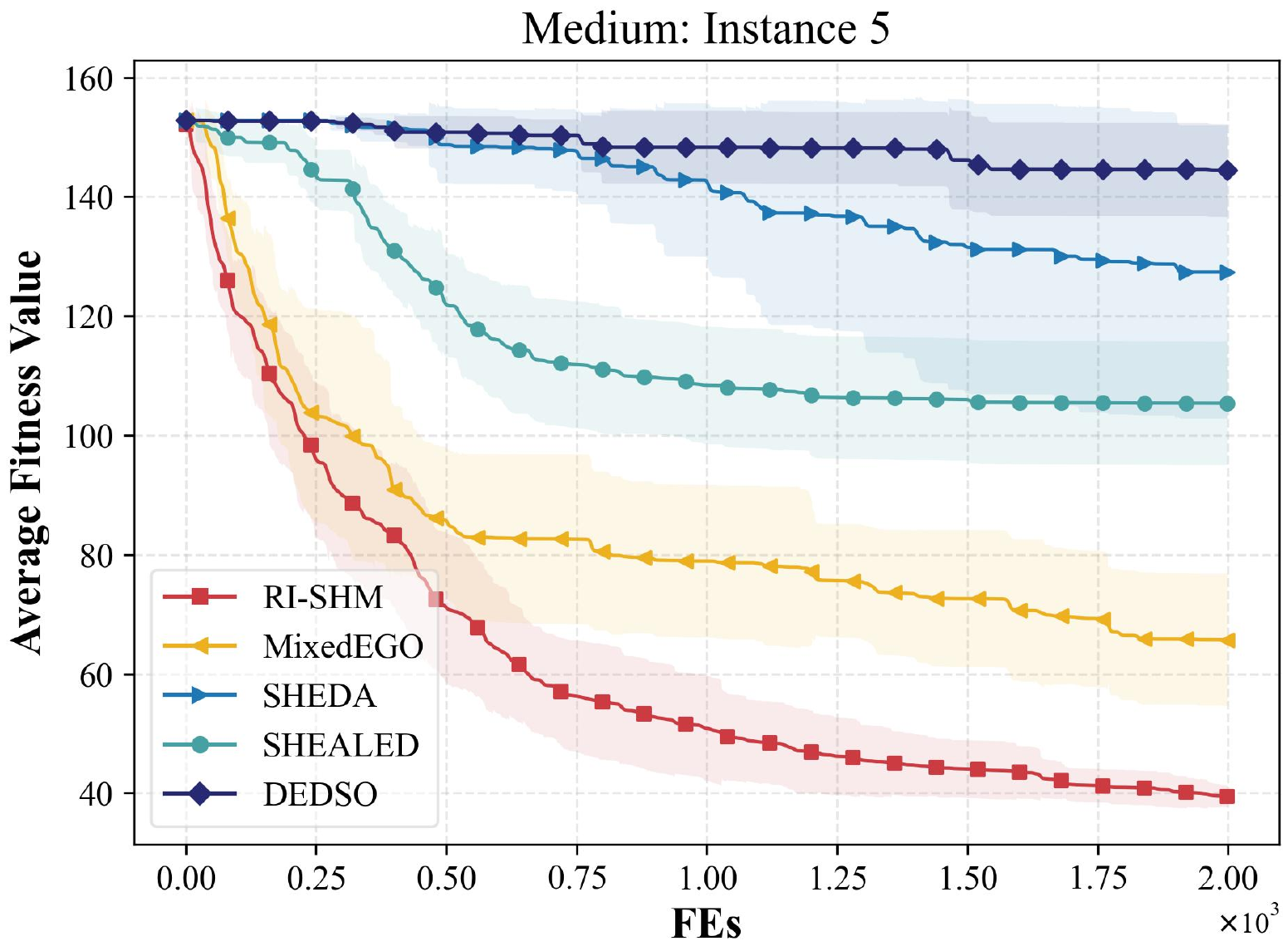}
       \label{fig:medium_05}}
    \\
    % 第一行子图
    \subfloat{%
       \includegraphics[width=0.195\textwidth]{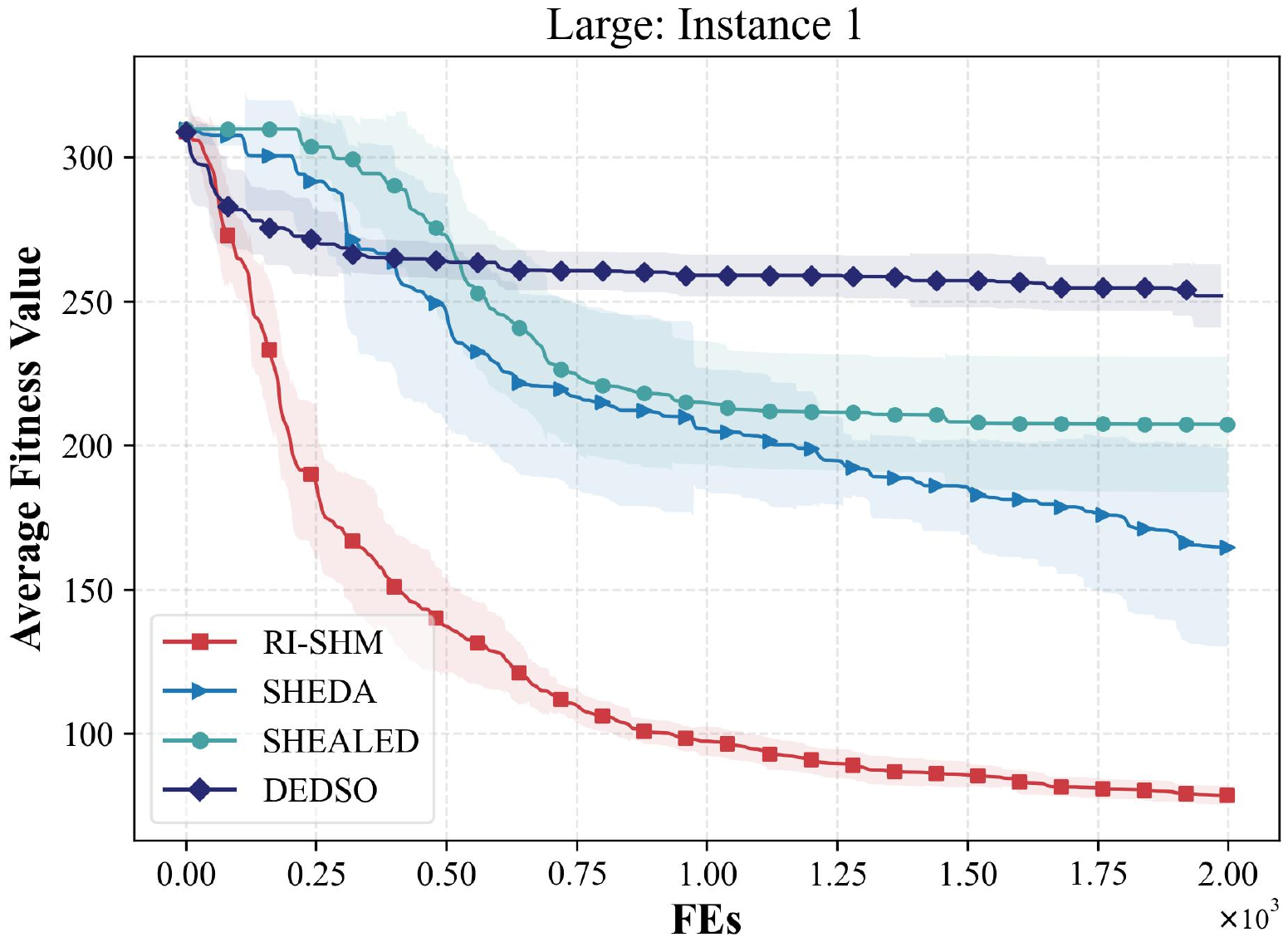}
       \label{fig:large_01}}
    \subfloat{%
       \includegraphics[width=0.195\textwidth]{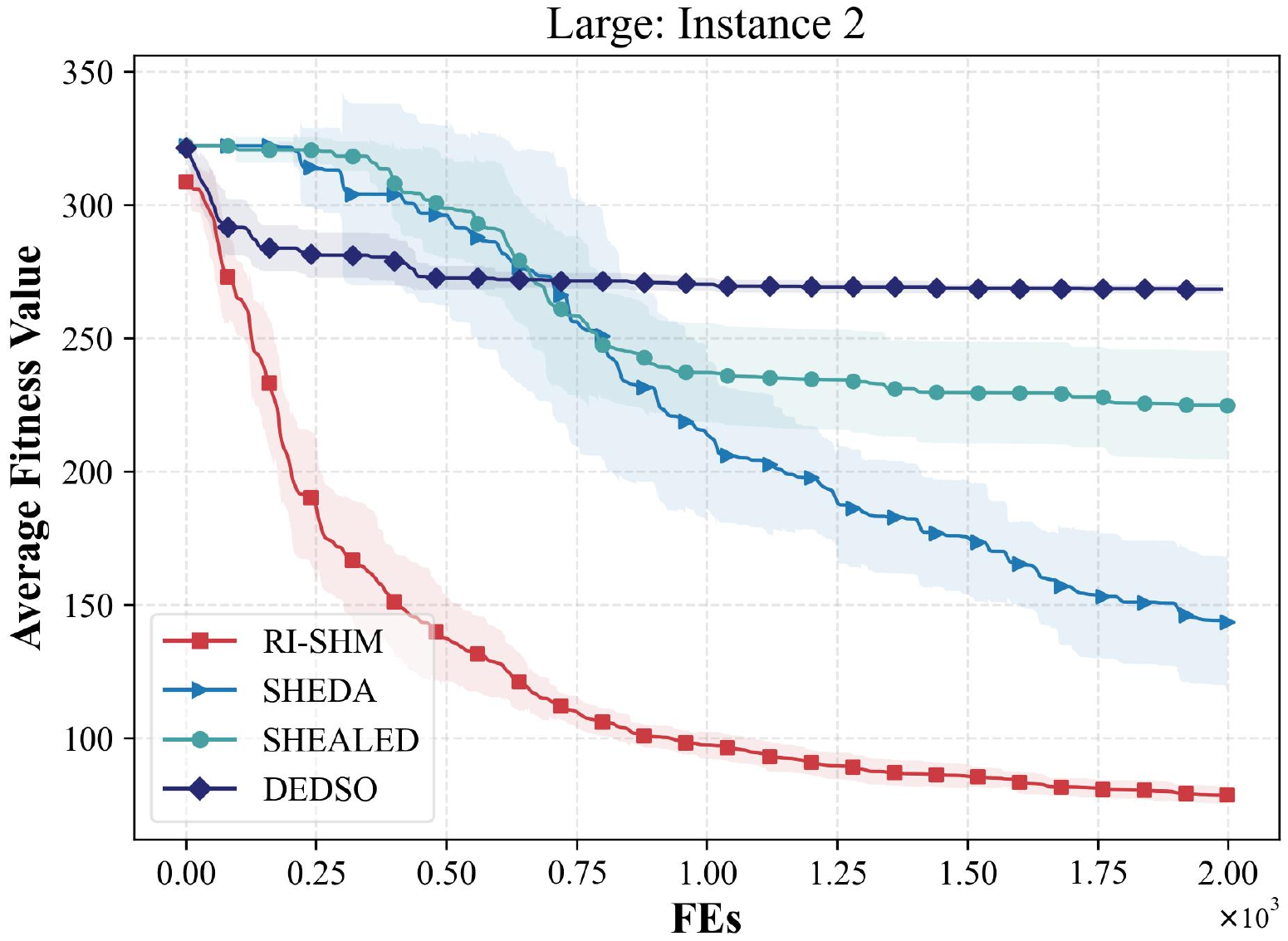}
       \label{fig:large_02}}
    \subfloat{%
       \includegraphics[width=0.195\textwidth]{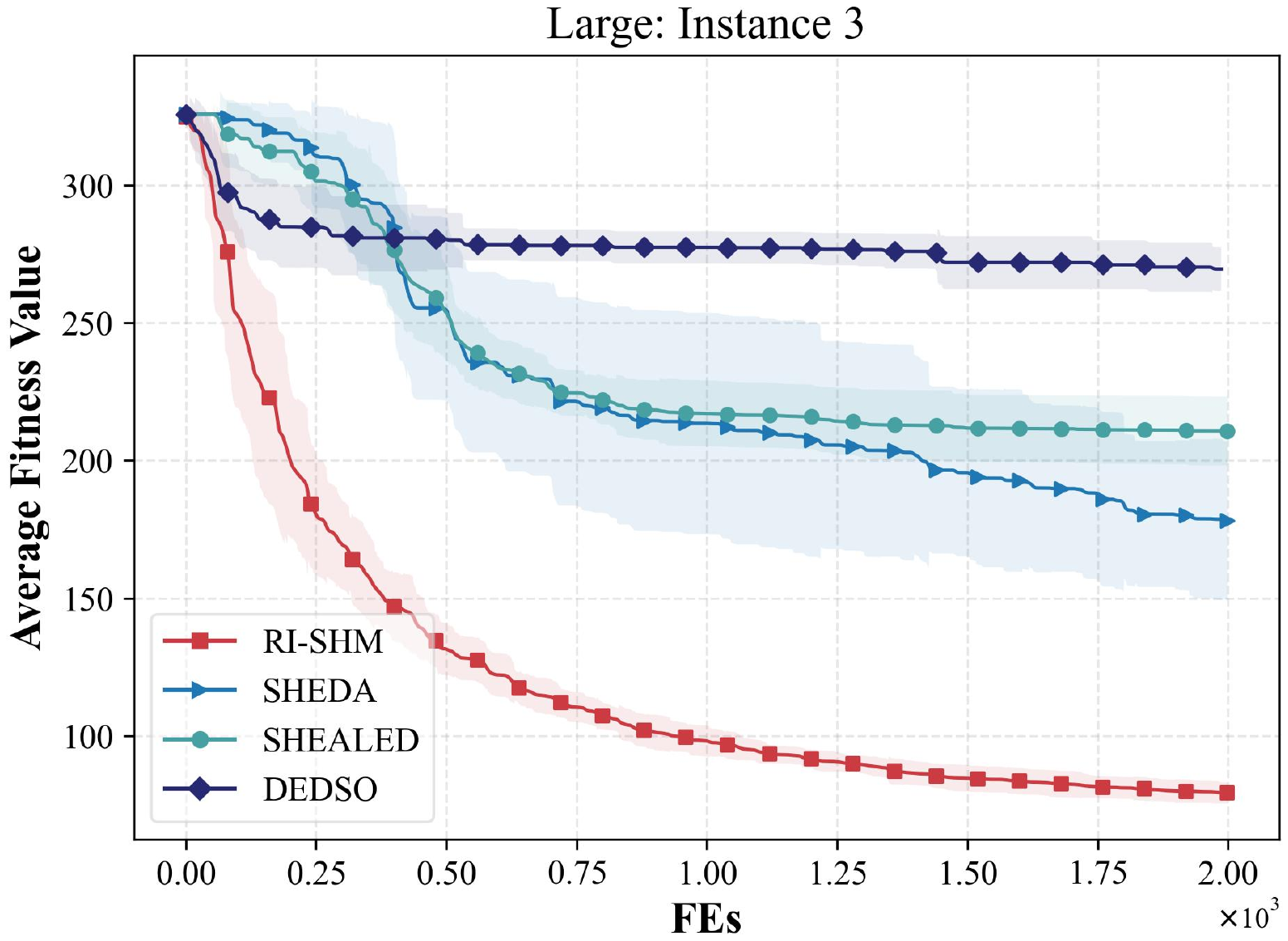}
       \label{fig:large_03}}
    \subfloat{%
       \includegraphics[width=0.195\textwidth]{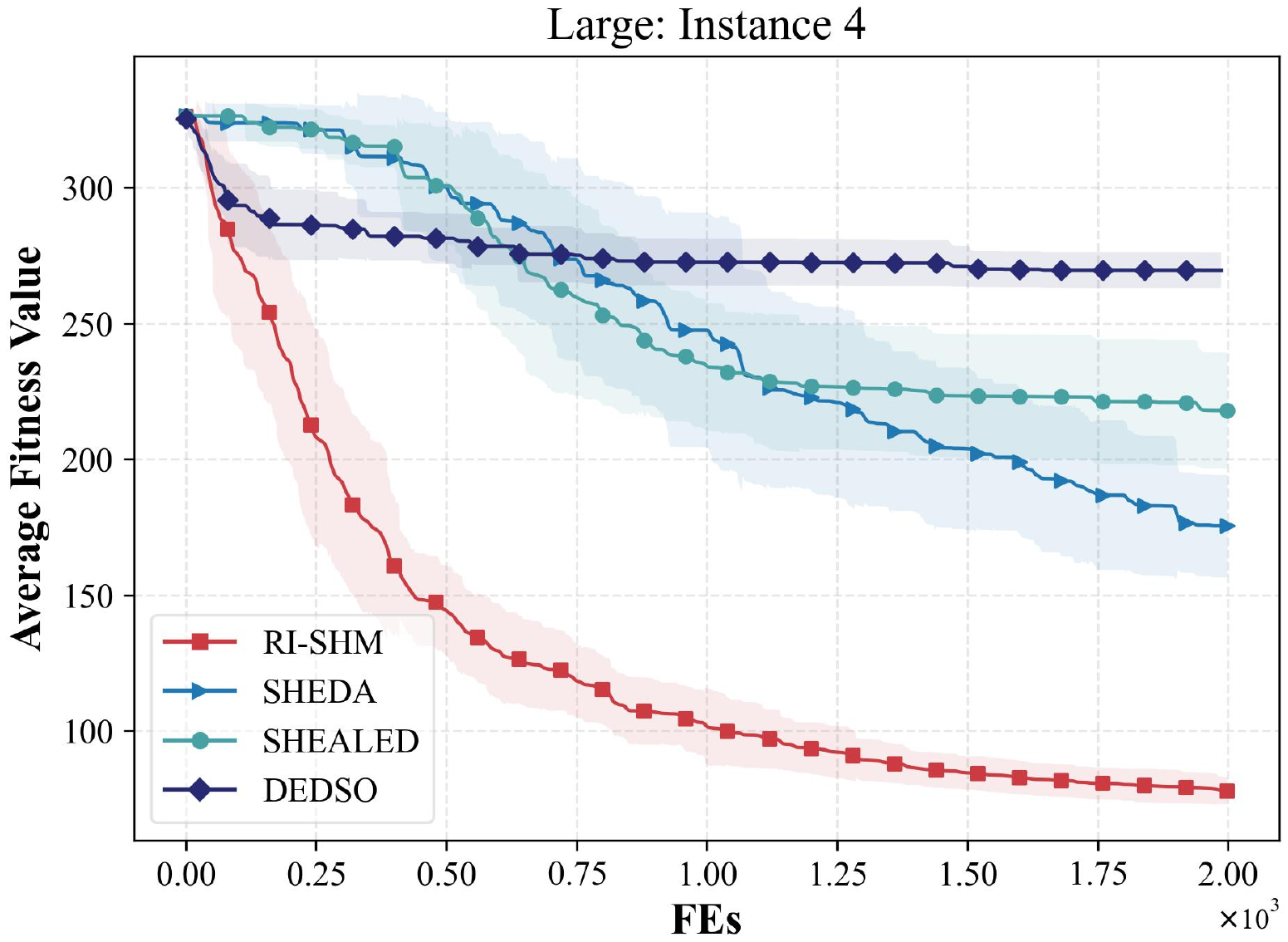}
       \label{fig:large_04}}
    \subfloat{%
       \includegraphics[width=0.195\textwidth]{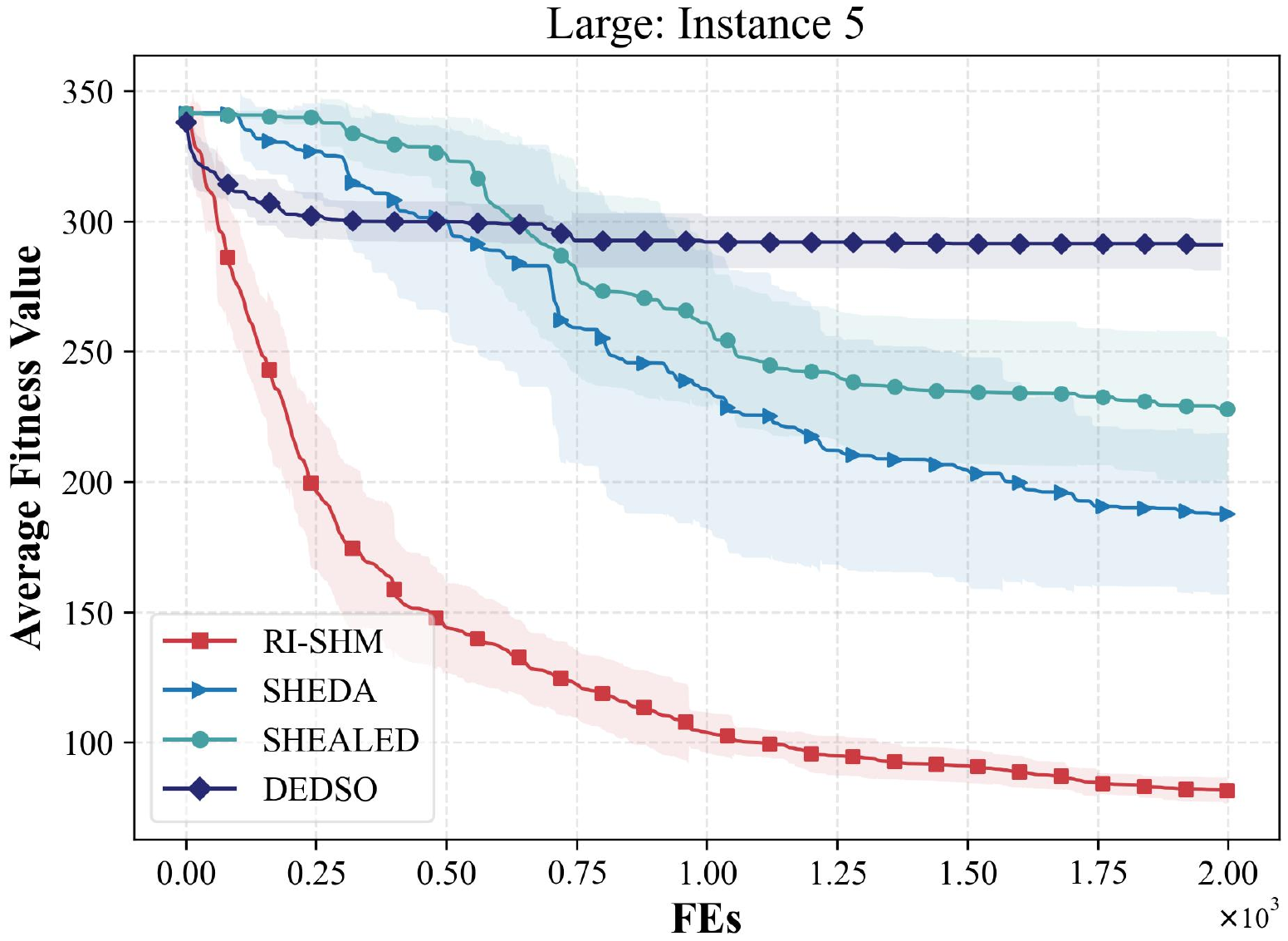}
       \label{fig:large_05}}
\caption{Convergence curves of the proposed RI-SHM and four comparative algorithms under different problem scales.}
\label{fig: convergence}
\end{figure*}

\subsection{Comparative Studies}
\subsubsection{Comparative Analysis of SAEAs}
\label{compare:rank}
Table \ref{compare:small} presents the performance comparison of RI-SHM with MixedEGO, SHEDA, SHEALED, and DEDSO across the problems with various scales. To assess statistical significance, we conduct the Wilcoxon rank sum test at a 5$\%$ significance level on fitness results. The symbols $+$, $\approx$, and $-$ denote that the corresponding algorithm performs significantly worse, comparably, or significantly better than RI-SHM, respectively.

As shown in Table \ref{compare:small}, RI-SHM consistently outperforms all comparative algorithms across the tested instances. Its advantages in stability and performance become more evident as the problem scale grows. Fig. \ref{fig: convergence} illustrates the convergence curves of comparative algorithms across 15 instances. 

In small and medium-scale scenarios, MixedEGO achieves results comparable to RI-SHM in certain instances, reflecting the strength of its kernel. However, RI-SHM exhibits a stronger ability to escape from local optima and continues to refine solutions during the later stages of evolution. 

SHEDA demonstrates rapid convergence within 1,000 FEs in small and medium-scale problems while failing to converge in large-scale scenarios, which indicates its strong exploratory potential. In contrast, DEDSO shows slower convergence in small and medium-scale problems and stagnates in large-scale scenarios. This indicates that DEDSO is more suitable for problems with fewer than 100 dimensions. 

In large-scale scenarios, RI-SHM stands out by efficiently exploring the solution space within limited FEs. This is benefited from the pairwise global surrogate's ability to identify promising solutions quickly. In comparison, algorithms like SHEALED perform well in small-scale problems and exhibit stagnation as problem complexity increases. These findings highlight RI-SHM’s scalability and effectiveness in addressing high-dimensional MVOPs.

% \usepackage{multirow}
\begin{table}[!t]
\centering
\scriptsize
\caption{Comparisons of RI-SHM and $CAGA_{mv}$ across all instances, presented as Avg(Std).}
\label{compare: original}
\begin{threeparttable}
\begin{tabular}{cc|c|c|c}
\toprule
\multicolumn{2}{c|}{Scale}                       & $CAGA_{mv}$-20K           & $CAGA_{mv}$-2K            & RI-SHM             \\ \midrule
\multicolumn{1}{c|}{\multirow{5}{*}{S}}  & 1 & 1.05E+01(2.21E-01) & 3.51E+01(2.20E+00) & 1.55E+01(1.67E+00) \\
\multicolumn{1}{c|}{}                        & 2 & 1.02E+01(4.31E-01) & 3.66E+01(3.46E+00) & 1.34E+01(1.52E+00) \\
\multicolumn{1}{c|}{}                        & 3 & 1.07E+01(6.34E-01) & 3.55E+01(1.91E+00) & 1.44E+01(1.74E+00) \\
\multicolumn{1}{c|}{}                        & 4 & 1.10E+01(2.99E-01) & 3.66E+01(1.58E+00) & 1.52E+01(2.32E+00) \\
\multicolumn{1}{c|}{}                        & 5 & 1.06E+01(3.90E-01) & 3.50E+01(2.16E+00) & 1.39E+01(1.01E+00) \\ \midrule
\multicolumn{1}{c|}{\multirow{5}{*}{M}} & 1 & 3.04E+01(8.34E-01) & 1.11E+02(5.34E+00) & 3.84E+01(2.45E+00) \\
\multicolumn{1}{c|}{}                        & 2 & 3.09E+01(7.66E-01) & 1.07E+02(8.06E+00) & 3.91E+01(3.14E+00) \\
\multicolumn{1}{c|}{}                        & 3 & 3.14E+01(1.34E+00) & 1.04E+02(8.39E+00) & 3.77E+01(1.30E+00) \\
\multicolumn{1}{c|}{}                        & 4 & 3.11E+01(8.60E-01) & 1.09E+02(9.12E+00) & 3.83E+01(3.32E+00) \\
\multicolumn{1}{c|}{}                        & 5 & 3.30E+01(1.65E+00) & 1.09E+02(5.03E+00) & 3.95E+01(1.81E+00) \\ \midrule
\multicolumn{1}{c|}{\multirow{5}{*}{L}}  & 1 & 6.76E+01(1.88E+00) & 2.45E+02(1.68E+01) & 7.86E+01(3.30E+00) \\
\multicolumn{1}{c|}{}                        & 2 & 6.55E+01(2.67E+00) & 2.41E+02(2.02E+01) & 7.71E+01(5.52E+00) \\
\multicolumn{1}{c|}{}                        & 3 & 6.67E+01(2.46E+00) & 2.57E+02(1.66E+01) & 7.92E+01(3.87E+00) \\
\multicolumn{1}{c|}{}                        & 4 & 6.57E+01(1.52E+00) & 2.49E+02(1.34E+01) & 77.9E+01(5.00E+00) \\
\multicolumn{1}{c|}{}                        & 5 & 7.11E+01(1.52E+00) & 2.51E+02(1.34E+01) & 8.14E+01(5.00E+00) \\ \bottomrule
\end{tabular}
\begin{tablenotes}    
                   %这行要添加
    \item \footnotesize $\bullet$ -2K indicates FEs are limited to 2000, consistent with RI-SHM, while -20K represents FEs are set to 20000.
\end{tablenotes} 
\end{threeparttable}
\vspace{-0.15cm}
\end{table}

\subsubsection{Comparisons of RI-SHM and $CAGA_{mv}$}
\begin{figure}[!t]
    \centering
    \includegraphics[width=0.95\linewidth]{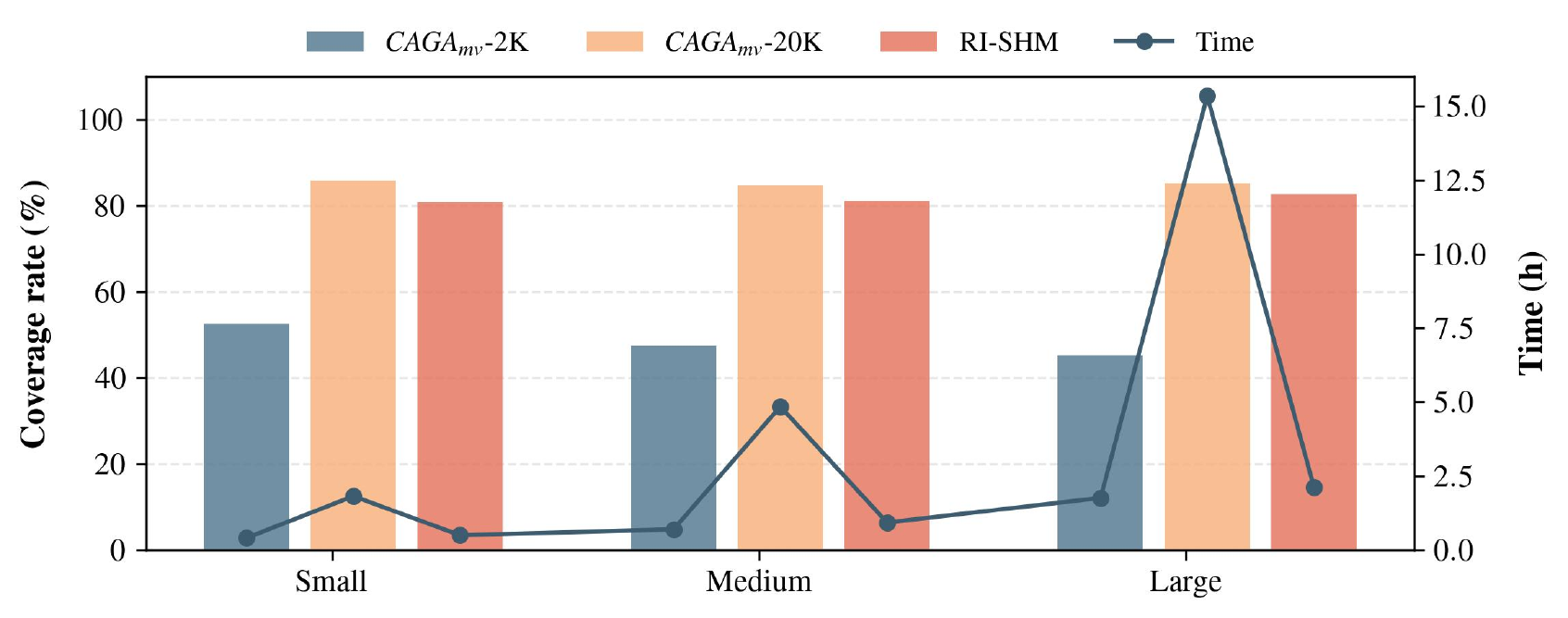}
    \caption{Dual-axis chart of average coverage rates and runtime for RI-SHM and $CAGA_{mv}$ across different problem scales.}
    \label{fig:original}
    \vspace{-0.15cm}
\end{figure}

We compare RI-SHM with two versions of $CAGA_{mv}$ \cite{wu2024mixed} in Table \ref{compare: original}. Under the same number of FEs, RI-SHM significantly outperforms $CAGA_{mv}$, highlighting the effectiveness of our SAEA framework. The smaller standard deviation indicates RI-SHM's ability to consistently achieve high-quality solutions by balancing global and local optimization phases. Furthermore, Fig. \ref{fig:original} presents that even with increasing problem complexity, RI-SHM achieves comparable coverage performance to $CAGA_{mv}$-20K with a slightly longer time. Here, the coverage rate is calculated as $Coverage=fitness/\sum_{q=1}^{|Q|}{\omega_{q}}$.

\subsubsection{Comparative Analysis of Runtime}
\begin{figure}[!t]
    \includegraphics[width=0.92\linewidth]{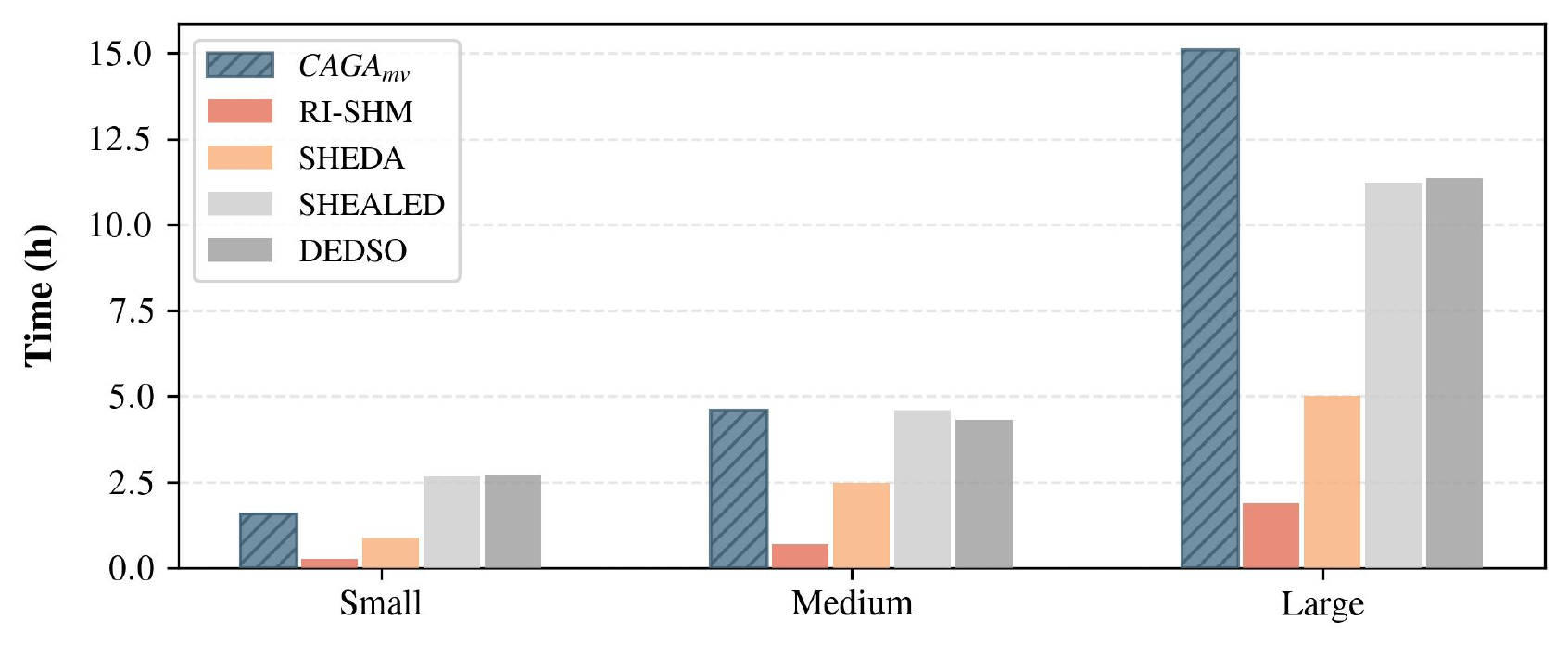}
    \caption{The comparison of average runtime for various SAEAs.}
    \label{fig:time}
\end{figure}

Our experiments also investigated the runtime of each algorithm. All algorithms are implemented in Python and executed under the same environment for fair comparison. MixedEGO is excluded due to its runtime exceeding 1 day across all problem scales. As shown in Fig. \ref{fig:time}, RI-SHM achieves the shortest runtime despite employing deep learning techniques. This efficiency is attributed to two aspects: (1) PyTorch’s training environment facilitates internal parallelization, and (2) the online update strategy in RI-SHM processes only a small number of new samples, reducing computational costs compared to full-data updates.

\subsection{Ablation Study}
To evaluate the contributions of each component in RI-SHM, we conducted an ablation study on this part. RI-SHM-w/oL and RI-SHM-w/oG are variants of our algorithm that exclude the local surrogate-assisted EDA and the global RankNet-based pairwise surrogate, respectively. This highlights the importance of the global and local modules. Specifically, one instance from each problem scale is randomly selected, and each variant is run 10 times independently. The statistical results are summarized in Table \ref{compare:ablation}.

\begin{figure}[!t]
\vspace{-0.15cm}
\centering
    % 第一行子图
    \subfloat[Medium-1]{%
       \includegraphics[width=0.49\linewidth]{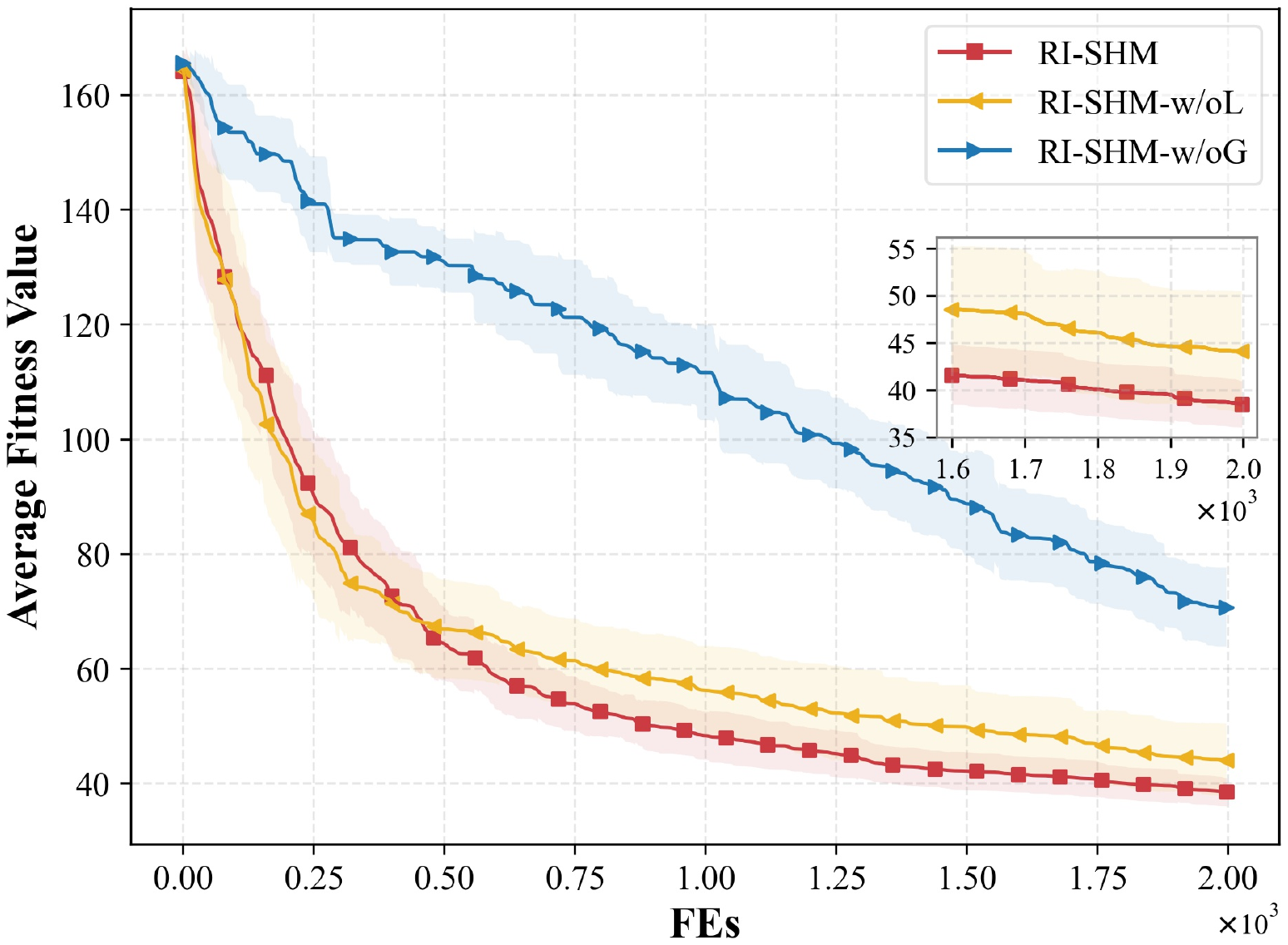}
       \label{fig:ablation_m}}
    \subfloat[Large-2]{%
       \includegraphics[width=0.49\linewidth]{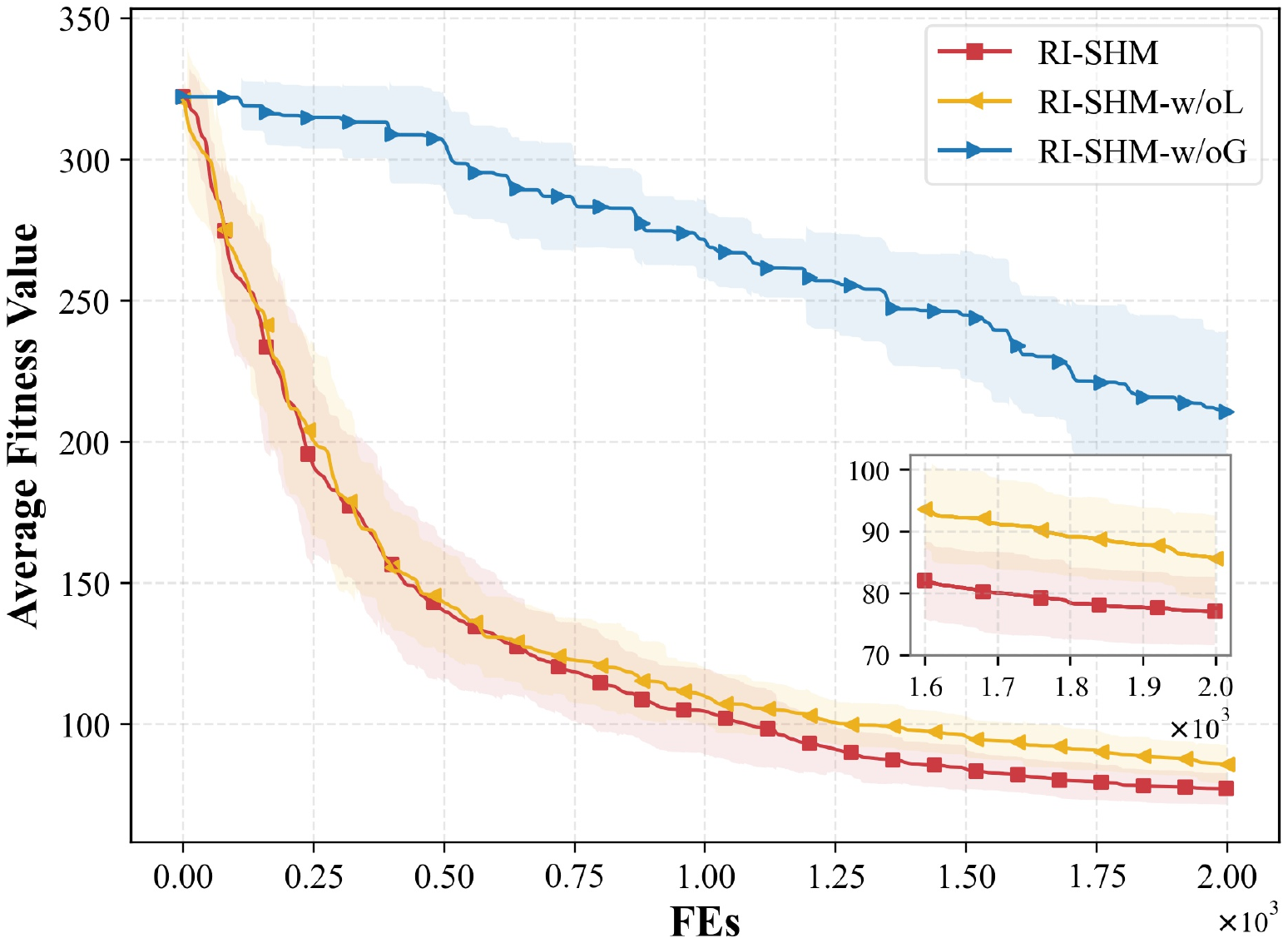}
       \label{fig:ablation_l}}
\caption{Convergence curves of RI-SHM's variants under different scales.}
\label{fig:ablation}
\end{figure}

Table \ref{compare:ablation} shows that RI-SHM significantly outperforms its variants, particularly RI-SHM-w/oG. Leveraging a small number of samples, the global surrogate enables fast online updates to explore the solution space and identify superior solutions. Fig. \ref{fig:ablation} presents the convergence curves for two instances of medium and large scales, further validating the global surrogate's effectiveness. The surrogate-assisted local EDA is complementary by helping the algorithm escape from local optima and refine existing solutions. Although its convergence speed is relatively slow, it is adequate for the demands of the local optimization phase. Without this component, achieving better coverage for sensor networks would be challenging. The global and local modules are essential to RI-SHM’s performance, as removing either leads to a notable decline in effectiveness.

\begin{table}[!t]
\setlength{\tabcolsep}{2pt}
\renewcommand{\arraystretch}{1.2}
\centering
\caption{Comparisons among different variants of the RI-SHM.}
\label{compare:ablation}
\scriptsize
\resizebox{\columnwidth}{!}{
\begin{tabular}{c|ccc|ccc|ccc}
\toprule
Instance && RI-SHM                      &&& RI-SHM-w/oG        &&& RI-SHM-w/oL       & \\ \midrule
S-5        && \cellcolor{gray!20}\textbf{1.39E+01(1.01E+00)} &&& 2.04E+01(3.88E+00) &&& 1.90E+01(3.29E+00) \\
M-1        && \cellcolor{gray!20}\textbf{3.84E+01(2.45E+00)} &&& 7.07E+01(6.88E+00) &&& 4.41E+01(6.38E+00) \\
L-2        && \cellcolor{gray!20}\textbf{7.71E+01(5.52E+00)} &&& 2.10E+02(2.84E+01) &&& 8.56E+01(6.90E+00) \\ \midrule
$+$/$\approx$/$-$            && -                           &&& 3 / 0 / 0              &&& 3 / 0 / 0              \\ \bottomrule
\end{tabular}}
\end{table}

\begin{table*}[!t]
\setlength{\tabcolsep}{2pt}
\renewcommand{\arraystretch}{1.2}
\centering
\caption{Comparisons of different surrogate models on predicting individual rankings, presented as Avg(Std). The best-performing results are highlighted.}
\label{tab:accuracy}
%\scriptsize
\resizebox{\textwidth}{!}{
\begin{tabular}{c|ccccc|ccccc|ccccc|ccccc}
\toprule
\multirow{2}{*}{Instance} & \multicolumn{5}{c|}{RankNet}      & \multicolumn{5}{c|}{RankSVM}      & \multicolumn{5}{c|}{RF}           & \multicolumn{5}{c}{RBFN}          \\
                          && Fitness            && Accuracy(\%) &&& Fitness            && Accuracy(\%) &&& Fitness            && Accuracy(\%) &&& Fitness            && Accuracy(\%) &\\ \midrule
S-2                       && \cellcolor{gray!20}\textbf{1.48E+01(1.40E+00)} && \cellcolor{gray!20}\textbf{65.27(2.05)} &&& 2.36E+01(3.56E+00) + && 58.09(2.71) + &&& 2.68E+01(5.40E+00) + && 64.20(2.63) +          &&& 2.19E+01(3.05E+00) + && 45.60(2.81) + \\
S-4                       && \cellcolor{gray!20}\textbf{1.63E+01(1.07E+00)} && 66.42(2.58)          &&& 2.05E+01(4.33E+00) + && 57.88(2.56) + &&& 3.24E+01(5.60E+00) + && \cellcolor{gray!20}\textbf{67.13(2.60) =} &&& 2.25E+01(1.55E+00) + && 46.45(5.03) + \\
M-1                       && \cellcolor{gray!20}\textbf{4.38E+01(5.98E+00)} && 64.20(0.87)          &&& 5.65E+01(9.02E+00) + && 59.14(1.77) + && &9.16E+01(2.20E+01) + && \cellcolor{gray!20}\textbf{65.96(2.52) -} &&& 6.42E+01(9.39E+00) + && 45.20(3.67) + \\
M-3                       && \cellcolor{gray!20}\textbf{4.07E+01(3.63E+00)} && 62.63(3.05)          &&& 6.37E+01(1.47E+01) + && 58.48(3.42) + &&& 6.37E+01(8.43E+00) + && \cellcolor{gray!20}\textbf{64.29(2.22) -} &&& 6.73E+01(1.09E+01) + && 43.76(3.43) + \\
L-2                       && \cellcolor{gray!20}\textbf{8.71E+01(6.82E+00)} && \cellcolor{gray!20}\textbf{62.37(1.08)} &&& 1.27E+02(3.45E+01) + && 60.01(2.11) + &&& 1.56E+02(3.21E+01) + && 61.68(2.24) =          &&& 1.38E+02(1.96E+01) + && 41.51(3.15) + \\
L-5                       && \cellcolor{gray!20}\textbf{8.81E+01(7.24E+00)} && \cellcolor{gray!20}\textbf{63.21(1.23)} &&& 1.29E+02(2.87E+01) + && 60.08(2.95) + &&& 1.68E+02(4.75E+01) + && 62.88(2.71) =          &&& 1.56E+02(2.49E+01) + && 38.87(5.92) + \\ 
\midrule
$+/\approx/-$                     & \multicolumn{5}{c|}{-}                             && 6 / 0 / 0               && 6 / 0 / 0        &&& 6 / 0 / 0               && 1 / 3 / 2                 &&& 6 / 0 / 0               && 6 / 0 / 0        &\\ \midrule
Average rank           & \multicolumn{5}{c|}{\textbf{1.25}}                          & \multicolumn{5}{c|}{2.33}          & \multicolumn{5}{c|}{2.67}                   & \multicolumn{5}{c}{3.58}           \\ \bottomrule
\end{tabular}}
\end{table*}

\subsection{Impact Analysis of RankNet-based Pairwise Surrogate}
\label{compare:impact}
To investigate the performance of the RankNet-based pairwise surrogate, we replace RI-SHM's global surrogate with RankSVM\footnote{https://gist.github.com/agramfort/2071994}\cite{lu2014new}, Random Forest (RF), and RBFN \cite{liu2023surrogate} for comparative analysis. The first two are pairwise-based, while the latter is regression-based. The surrogate-assisted local EDA was removed to assess the global surrogate's influence. We follow the method described in \cite{tian2023pairwise} to construct training and test datasets. The initial archive is used as training data to ensure uniform training conditions across surrogate models. In contrast, offspring solutions generated by the population are used as the test set to evaluate the model's performance in predicting the ranking between pairs of individuals. Fitness performance and prediction accuracy are used as evaluation metrics. For all the compared learning algorithms, no hyperparameter optimization is conducted for fair comparison.

Table \ref{tab:accuracy} presents the statistical performance and prediction accuracy across six instances with varying scales, where "RankNet" refers to the RankNet-based pairwise global surrogate. We also conduct the Wilcoxon rank-sum test similar to Section \ref{compare:rank} on the results and calculate the average rankings with the Friedman test. The analysis reveals that RankNet achieves better rankings than other algorithms. It attains the best statistical results and outperforms RankSVM and RBFN in accuracy. RF delivers comparable predictive accuracy for individual rankings and even outperforms RankNet in some instances due to its ensemble learning mechanism. However, RF-based optimization shows poor fitness performance, as it can identify superior individuals but struggles to capture significant changes, often leading to stagnation in local optima. RBFN performs the worst overall, highlighting the advantages of pairwise-based surrogates over regression-based ones.

\begin{figure}
\vspace{-0.2cm}
    \centering
    \includegraphics[width=0.9\linewidth]{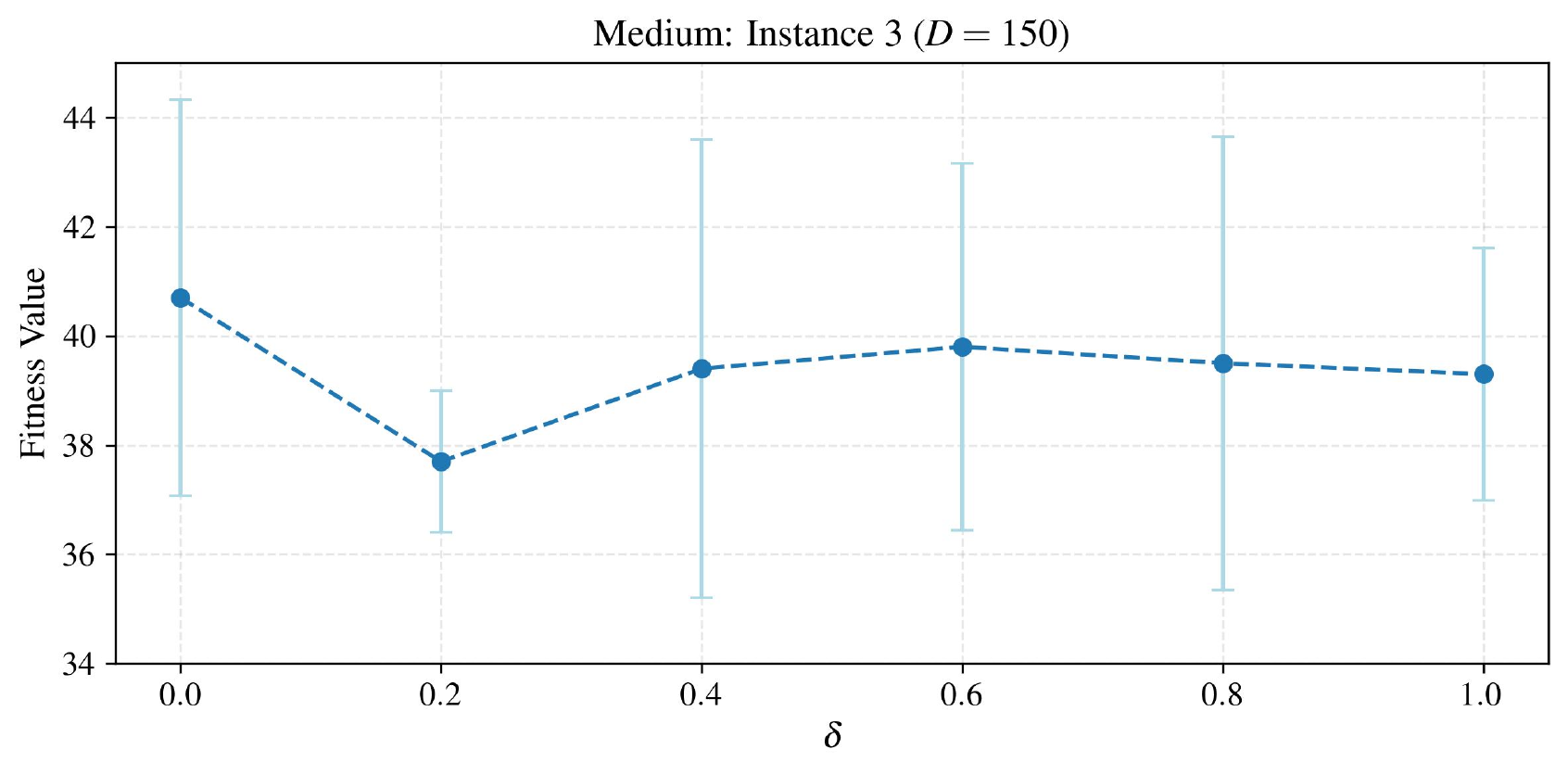}
    \caption{Fitness value obtained by RI-SHM with different threshold $\delta$.}
    \label{fig:para}
\end{figure}
\subsection{Parameter Sensitivity Analysis}
The threshold $\delta$ is a core component of RI-SHM, which controls the switching between global and local optimization phases to balance exploration and exploitation. At the end of each iteration, the algorithm evaluates the fitness diversity of the current population. If the fitness diversity falls below $\delta$, the algorithm switches to the alternate optimizer to detect fresh genotypes. To determine the appropriate value for $\delta$, we randomly select a medium-scale instance as the test scenario. The medium-scale instance represents a balanced compromise between small and large-scale cases, making it suitable for evaluating performance across varying problem scales. 

Fig. \ref{fig:para} illustrates the fitness performance of RI-SHM under different threshold values of $\delta$. The results indicate that $\delta= 0.2$ achieves the best average fitness value and standard deviation, making it the suitable choice for all experiments in this study. When $\delta$ equals 0, RI-SHM relies exclusively on the global surrogate for exploration. Conversely, when $\delta$ equals 1, RI-SHM continuously alternates between the global and local optimizers at each iteration. The suboptimal performance observed in both cases highlights the importance of switching appropriately between the global and local optimizers. Additionally, the difference between the best and worst average fitness values under various $\delta$ is about 7.5$\%$, which means that RI-SHM is not highly sensitive to $\delta$. 

\section{Conclusion and future work}
\label{sec:conclude}
In this paper, We reformulate coverage optimization via the lens of
EOPs to create a high-fidelity realistic model, and propose a novel RI-SHM to address this EMVOP. It generally integrates three components. First, the RankNet-based pairwise global surrogate predicts the rankings between pairs of individuals, which enables the identification of promising individuals for real FEs. Second, the surrogate-assisted local EDA is employed to continuously refine solutions and help escape from local optima. Finally, the fitness diversity-driven strategy adaptively switches between global and local
optimizers during the evolution.

To validate the effectiveness of RI-SHM, we conduct experiments on 15 test instances across different problem scales, with dimensions ranging from 75 to 300. Comparative results against other algorithms demonstrate the superiority of RI-SHM in terms of both solution quality and robustness. Additionally, we further investigate the algorithm's components and find that the predictive accuracy of RankNet-based surrogate is usually superior to other models. These findings underscore the capability of RI-SHM to address high-dimensional EMVOPs effectively. In the future, we aim to extend RI-SHM to standard benchmark Black-Box Optimization problems and address a broader range of typical EMVOPs (i.e., NAS). Additionally, exploring coverage optimization under uncertainty is another promising direction \cite{xiang2019clustering}.

% \begin{thebibliography}{99}
\bibliographystyle{IEEEtran}
\bibliography{IEEEabrv, mylib}

\begin{thebibliography}{10}
\providecommand{\url}[1]{#1}
\csname url@samestyle\endcsname
\providecommand{\newblock}{\relax}
\providecommand{\bibinfo}[2]{#2}
\providecommand{\BIBentrySTDinterwordspacing}{\spaceskip=0pt\relax}
\providecommand{\BIBentryALTinterwordstretchfactor}{4}
\providecommand{\BIBentryALTinterwordspacing}{\spaceskip=\fontdimen2\font plus
\BIBentryALTinterwordstretchfactor\fontdimen3\font minus \fontdimen4\font\relax}
\providecommand{\BIBforeignlanguage}[2]{{%
\expandafter\ifx\csname l@#1\endcsname\relax
\typeout{** WARNING: IEEEtran.bst: No hyphenation pattern has been}%
\typeout{** loaded for the language `#1'. Using the pattern for}%
\typeout{** the default language instead.}%
\else
\language=\csname l@#1\endcsname
\fi
#2}}
\providecommand{\BIBdecl}{\relax}
\BIBdecl

\bibitem{laporte2019introduction}
G.~Laporte, S.~Nickel, and F.~Saldanha-da Gama, \emph{Introduction to Location Science}.\hskip 1em plus 0.5em minus 0.4em\relax Springer, 2019.

\bibitem{ding2021metaheuristics}
S.~Ding, C.~Chen, Q.~Zhang, B.~Xin, and P.~Pardalos, \emph{Metaheuristics for Resource Deployment under Uncertainty in Complex Systems}.\hskip 1em plus 0.5em minus 0.4em\relax CRC Press, 2021.

\bibitem{ma2023service}
X.~Ma, X.~Zhang, and X.~Zhao, ``Service coverage optimization for facility location: considering line-of-sight coverage in continuous demand space,'' \emph{International Journal of Geographical Information Science}, vol.~37, no.~7, pp. 1496--1519, 2023.

\bibitem{yuan2024joint}
X.~Yuan, F.~Tang, M.~Zhao, and N.~Kato, ``Joint rate and coverage optimization for the thz/rf multi-band communications of space-air-ground integrated network in 6g,'' \emph{IEEE Transactions on Wireless Communications}, vol.~23, no.~6, pp. 6669--6682, 2024.

\bibitem{he2022collaborative}
S.~He, K.~Shi, C.~Liu, B.~Guo, J.~Chen, and Z.~Shi, ``Collaborative sensing in internet of things: A comprehensive survey,'' \emph{IEEE Communications Surveys \& Tutorials}, vol.~24, no.~3, pp. 1435--1474, 2022.

\bibitem{cao2018deployment}
B.~Cao, J.~Zhao, Z.~Lv, X.~Liu, X.~Kang, and S.~Yang, ``Deployment optimization for 3d industrial wireless sensor networks based on particle swarm optimizers with distributed parallelism,'' \emph{Journal of Network and Computer Applications}, vol. 103, pp. 225--238, 2018.

\bibitem{saad2020toward}
A.~Saad, M.~R. Senouci, and O.~Benyattou, ``Toward a realistic approach for the deployment of 3d wireless sensor networks,'' \emph{IEEE Transactions on Mobile Computing}, vol.~21, no.~4, pp. 1508--1519, 2020.

\bibitem{wu2024mixed}
T.~Wu, Y.~Zhang, C.~Miao, C.~Chen, and S.~Ding, ``Mixed-variable correlation-aware metaheuristic for deployment optimization of 3-d sensor networks,'' in \emph{Proceedings of the Genetic and Evolutionary Computation Conference}, 2024, pp. 1390--1398.

\bibitem{heyns2021optimisation}
A.~M. Heyns, ``Optimisation of surveillance camera site locations and viewing angles using a novel multi-attribute, multi-objective genetic algorithm: A day/night anti-poaching application,'' \emph{Computers, Environment and Urban Systems}, vol.~88, p. 101638, 2021.

\bibitem{zhu2023maximal}
X.~Zhu and M.~Zhou, ``Maximal weighted coverage deployment of uav-enabled rechargeable visual sensor networks,'' \emph{IEEE Transactions on Intelligent Transportation Systems}, vol.~24, no.~10, pp. 11\,293--11\,307, 2023.

\bibitem{liu2023surrogate}
Y.~Liu and H.~Wang, ``Surrogate-assisted hybrid evolutionary algorithm with local estimation of distribution for expensive mixed-variable optimization problems,'' \emph{Applied Soft Computing}, vol. 133, p. 109957, 2023.

\bibitem{cordeau2019benders}
J.-F. Cordeau, F.~Furini, and I.~Ljubi{\'c}, ``Benders decomposition for very large scale partial set covering and maximal covering location problems,'' \emph{European Journal of Operational Research}, vol. 275, no.~3, pp. 882--896, 2019.

\bibitem{zhang2024acu}
Y.~Zhang, C.~Chen, S.~Ding, and F.~Deng, ``An integer programming approach for angular coverage under uncertainty,'' in \emph{IEEE Conference on Decision and Control}.\hskip 1em plus 0.5em minus 0.4em\relax IEEE, 2024, pp. 1--6.

\bibitem{yao2019location}
J.~Yao, X.~Zhang, and A.~T. Murray, ``Location optimization of urban fire stations: Access and service coverage,'' \emph{Computers, Environment and Urban Systems}, vol.~73, pp. 184--190, 2019.

\bibitem{wrozynski2024reaching}
R.~Wr{\'o}{\.z}y{\'n}ski, K.~Pyszny, and M.~Wr{\'o}{\.z}y{\'n}ska, ``Reaching beyond gis for comprehensive 3d visibility analysis,'' \emph{Landscape and Urban Planning}, vol. 247, p. 105074, 2024.

\bibitem{liao2013ant}
T.~Liao, K.~Socha, M.~A.~M. de~Oca, T.~St{\"u}tzle, and M.~Dorigo, ``Ant colony optimization for mixed-variable optimization problems,'' \emph{IEEE Transactions on Evolutionary Computation}, vol.~18, no.~4, pp. 503--518, 2013.

\bibitem{talbi2009metaheuristics}
E.-G. Talbi, \emph{{Metaheuristics: from Design to Implementation}}.\hskip 1em plus 0.5em minus 0.4em\relax John Wiley \& Sons, 2009.

\bibitem{li2021surrogate}
J.-Y. Li, Z.-H. Zhan, J.~Xu, S.~Kwong, and J.~Zhang, ``Surrogate-assisted hybrid-model estimation of distribution algorithm for mixed-variable hyperparameters optimization in convolutional neural networks,'' \emph{IEEE Transactions on Neural Networks and Learning Systems}, vol.~34, no.~5, pp. 2338--2352, 2021.

\bibitem{jones1998efficient}
D.~R. Jones, M.~Schonlau, and W.~J. Welch, ``Efficient global optimization of expensive black-box functions,'' \emph{Journal of Global optimization}, vol.~13, pp. 455--492, 1998.

\bibitem{wang2017committee}
H.~Wang, Y.~Jin, and J.~Doherty, ``Committee-based active learning for surrogate-assisted particle swarm optimization of expensive problems,'' \emph{IEEE Transactions on Cybernetics}, vol.~47, no.~9, pp. 2664--2677, 2017.

\bibitem{song2021kriging}
Z.~Song, H.~Wang, C.~He, and Y.~Jin, ``A kriging-assisted two-archive evolutionary algorithm for expensive many-objective optimization,'' \emph{IEEE Transactions on Evolutionary Computation}, vol.~25, no.~6, pp. 1013--1027, 2021.

\bibitem{xie2023dual}
L.~Xie, G.~Li, K.~Lin, and Z.~Wang, ``Dual-state-driven evolutionary optimization for expensive optimization problems with continuous and categorical variables,'' in \emph{2023 5th International Conference on Data-driven Optimization of Complex Systems (DOCS)}.\hskip 1em plus 0.5em minus 0.4em\relax IEEE, 2023, pp. 1--7.

\bibitem{saves2024smt}
P.~Saves, R.~Lafage, N.~Bartoli, Y.~Diouane, J.~Bussemaker, T.~Lefebvre, J.~T. Hwang, J.~Morlier, and J.~R. Martins, ``Smt 2.0: A surrogate modeling toolbox with a focus on hierarchical and mixed variables gaussian processes,'' \emph{Advances in Engineering Software}, vol. 188, p. 103571, 2024.

\bibitem{hao2022expensive}
H.~Hao, A.~Zhou, H.~Qian, and H.~Zhang, ``Expensive multiobjective optimization by relation learning and prediction,'' \emph{IEEE Transactions on Evolutionary Computation}, vol.~26, no.~5, pp. 1157--1170, 2022.

\bibitem{liu2013gaussian}
B.~Liu, Q.~Zhang, and G.~G. Gielen, ``A gaussian process surrogate model assisted evolutionary algorithm for medium scale expensive optimization problems,'' \emph{IEEE Transactions on Evolutionary Computation}, vol.~18, no.~2, pp. 180--192, 2013.

\bibitem{yuan2021expensive}
Y.~Yuan and W.~Banzhaf, ``Expensive multiobjective evolutionary optimization assisted by dominance prediction,'' \emph{IEEE Transactions on Evolutionary Computation}, vol.~26, no.~1, pp. 159--173, 2021.

\bibitem{tian2023pairwise}
Y.~Tian, J.~Hu, C.~He, H.~Ma, L.~Zhang, and X.~Zhang, ``A pairwise comparison based surrogate-assisted evolutionary algorithm for expensive multi-objective optimization,'' \emph{Swarm and Evolutionary Computation}, vol.~80, p. 101323, 2023.

\bibitem{burges2005learning}
C.~Burges, T.~Shaked, E.~Renshaw, A.~Lazier, M.~Deeds, N.~Hamilton, and G.~Hullender, ``Learning to rank using gradient descent,'' in \emph{Proceedings of the 22nd International Conference on Machine Learning}, 2005, pp. 89--96.

\bibitem{zhang2024two}
Z.~Zhang, Y.~Wang, J.~Liu, G.~Sun, and K.~Tang, ``A two-phase kriging-assisted evolutionary algorithm for expensive constrained multiobjective optimization problems,'' \emph{IEEE Transactions on Systems, Man, and Cybernetics: Systems}, 2024.

\bibitem{lu2014new}
X.~Lu, K.~Tang, B.~Sendhoff, and X.~Yao, ``A new self-adaptation scheme for differential evolution,'' \emph{Neurocomputing}, vol. 146, pp. 2--16, 2014.

\bibitem{zhang2015classification}
J.~Zhang, A.~Zhou, and G.~Zhang, ``A classification and pareto domination based multiobjective evolutionary algorithm,'' in \emph{2015 IEEE Congress on Evolutionary Computation (CEC)}.\hskip 1em plus 0.5em minus 0.4em\relax IEEE, 2015, pp. 2883--2890.

\bibitem{pan2018classification}
L.~Pan, C.~He, Y.~Tian, H.~Wang, X.~Zhang, and Y.~Jin, ``A classification-based surrogate-assisted evolutionary algorithm for expensive many-objective optimization,'' \emph{IEEE Transactions on Evolutionary Computation}, vol.~23, no.~1, pp. 74--88, 2018.

\bibitem{dushatskiy2019convolutional}
A.~Dushatskiy, A.~M. Mendrik, T.~Alderliesten, and P.~A. Bosman, ``Convolutional neural network surrogate-assisted gomea,'' in \emph{Proceedings of the Genetic and Evolutionary Computation Conference}, 2019, pp. 753--761.

\bibitem{deif2013classification}
D.~S. Deif and Y.~Gadallah, ``Classification of wireless sensor networks deployment techniques,'' \emph{IEEE Communications Surveys \& Tutorials}, vol.~16, no.~2, pp. 834--855, 2013.

\bibitem{feng2021unknown}
S.~Feng, H.~Shi, L.~Huang, S.~Shen, S.~Yu, H.~Peng, and C.~Wu, ``Unknown hostile environment-oriented autonomous wsn deployment using a mobile robot,'' \emph{Journal of Network and Computer Applications}, vol. 182, p. 103053, 2021.

\bibitem{zhu2023multiobjective}
X.~Zhu and M.~Zhou, ``Multiobjective optimized deployment of edge-enabled wireless visual sensor networks for target coverage,'' \emph{IEEE Internet of Things Journal}, vol.~10, no.~17, pp. 15\,325--15\,337, 2023.

\bibitem{nguyen2018efficient}
L.~V. Nguyen, G.~Hu, and C.~J. Spanos, ``Efficient sensor deployments for spatio-temporal environmental monitoring,'' \emph{IEEE Transactions on Systems, Man, and Cybernetics: Systems}, vol.~50, no.~12, pp. 5306--5316, 2018.

\bibitem{xiang2019clustering}
X.~Xiang, Y.~Tian, J.~Xiao, and X.~Zhang, ``A clustering-based surrogate-assisted multiobjective evolutionary algorithm for shelter location problem under uncertainty of road networks,'' \emph{IEEE Transactions on Industrial Informatics}, vol.~16, no.~12, pp. 7544--7555, 2019.

\bibitem{jin2011surrogate}
Y.~Jin, ``Surrogate-assisted evolutionary computation: Recent advances and future challenges,'' \emph{Swarm and Evolutionary Computation}, vol.~1, no.~2, pp. 61--70, 2011.

\bibitem{zhou2024evolutionary}
M.~Zhou, M.~Cui, D.~Xu, S.~Zhu, Z.~Zhao, and A.~Abusorrah, ``Evolutionary optimization methods for high-dimensional expensive problems: A survey,'' \emph{IEEE/CAA Journal of Automatica Sinica}, vol.~11, no.~5, pp. 1092--1105, 2024.

\bibitem{wei2023hybrid}
F.-F. Wei, W.-N. Chen, and J.~Zhang, ``A hybrid regressor and classifier-assisted evolutionary algorithm for expensive optimization with incomplete constraint information,'' \emph{IEEE Transactions on Systems, Man, and Cybernetics: Systems}, vol.~53, no.~8, pp. 5071--5083, 2023.

\bibitem{temel2013deployment}
S.~Temel, N.~Unaldi, and O.~Kaynak, ``On deployment of wireless sensors on 3-d terrains to maximize sensing coverage by utilizing cat swarm optimization with wavelet transform,'' \emph{IEEE Transactions on Systems, Man, and Cybernetics: Systems}, vol.~44, no.~1, pp. 111--120, 2013.

\bibitem{wang2019novel}
X.~Wang, G.~G. Wang, B.~Song, P.~Wang, and Y.~Wang, ``A novel evolutionary sampling assisted optimization method for high-dimensional expensive problems,'' \emph{IEEE Transactions on Evolutionary Computation}, vol.~23, no.~5, pp. 815--827, 2019.

\bibitem{bradley1952rank}
R.~A. Bradley and M.~E. Terry, ``Rank analysis of incomplete block designs: I. the method of paired comparisons,'' \emph{Biometrika}, vol.~39, no. 3/4, pp. 324--345, 1952.

\bibitem{vaswani2017attention}
A.~Vaswani, N.~Shazeer, N.~Parmar, J.~Uszkoreit, L.~Jones, A.~N. Gomez, L.~u. Kaiser, and I.~Polosukhin, ``Attention is all you need,'' in \emph{Advances in Neural Information Processing Systems}, vol.~30, 2017, pp. 5998--6008.

\bibitem{van2001art}
D.~A. Van~Dyk and X.-L. Meng, ``The art of data augmentation,'' \emph{Journal of Computational and Graphical Statistics}, vol.~10, no.~1, pp. 1--50, 2001.

\bibitem{polakova2019differential}
R.~Pol{\'a}kov{\'a}, J.~Tvrd{\'\i}k, and P.~Bujok, ``Differential evolution with adaptive mechanism of population size according to current population diversity,'' \emph{Swarm and Evolutionary Computation}, vol.~50, p. 100519, 2019.

\bibitem{neri2012memetic}
F.~Neri and C.~Cotta, ``Memetic algorithms and memetic computing optimization: A literature review,'' \emph{Swarm and Evolutionary Computation}, vol.~2, pp. 1--14, 2012.

\bibitem{lian2012three}
X.-y. Lian, J.~Zhang, C.~Chen, and F.~Deng, ``Three-dimensional deployment optimization of sensor network based on an improved particle swarm optimization algorithm,'' in \emph{Proceedings of the 10th World Congress on Intelligent Control and Automation}.\hskip 1em plus 0.5em minus 0.4em\relax IEEE, 2012, pp. 4395--4400.

\bibitem{zhang2023surrogate}
Y.~Zhang, C.~Chen, T.~Wu, C.~Miao, and S.~Ding, ``Surrogate-assisted hybrid metaheuristic for mixed-variable 3-d deployment optimization of directional sensor networks,'' in \emph{2023 5th International Conference on Data-driven Optimization of Complex Systems (DOCS)}.\hskip 1em plus 0.5em minus 0.4em\relax IEEE, 2023, pp. 1--9.

\end{thebibliography}

% \end{thebibliography}

% \newpage

 

% \end{IEEEbiography}



% \vfill

\end{document}